\documentclass[11pt]{article}
\usepackage[letterpaper,margin=1in]{geometry}
\usepackage[numbers,sort&compress]{natbib}
\usepackage[T1]{fontenc}
\usepackage{mathptmx}
\usepackage{amsmath,amssymb}
\usepackage{booktabs}
\usepackage{graphicx}
\usepackage{microtype}
\usepackage{url}
\usepackage{float}
\usepackage[hidelinks]{hyperref}
\input{numbers}
\input{numbers2}
\input{numbers3}

\hypersetup{%
  pdftitle={Replanning in Planar Pushing with Learned Dynamics: Frequency, Timing, and Search Effort},
  pdfauthor={Shengwei Zhang, Haohua Liu, Shenkang Qian, Mo Shi}
}

\title{Replanning in Planar Pushing with Learned Dynamics:\\Frequency, Timing, and Search Effort}
\author{%
  Shengwei Zhang\textsuperscript{1,\ensuremath{\dagger}}\quad
  Haohua Liu\textsuperscript{2,\ensuremath{\dagger}}\quad
  Shenkang Qian\textsuperscript{3,\ensuremath{\dagger}}\quad
  Mo Shi\textsuperscript{3,\ensuremath{\dagger}}\\[0.55em]
  {\normalsize\textsuperscript{1}University of Pennsylvania\quad
  \textsuperscript{2}Imperial College London}\\
  {\normalsize\textsuperscript{3}University College London}\\[0.4em]
  {\small\textsuperscript{\ensuremath{\dagger}}\,These authors contributed equally and share first authorship.}%
}
\date{}

\begin{document}
\maketitle

\begin{abstract}
Choosing when to replan trades off access to new observations against search cost. We compare four signal-triggered rules on simulated planar pushing (PushT) using a curve of fixed periods and two controls: random replans matched to each rule's overall rate, and schedules shuffled across tasks to preserve the rule's replan frequency at each episode step. Across \S{st.seeds} state-input training seeds and 64 tasks, replanning every step succeeded on \S{st.fixed1.mean} tasks on average versus \S{st.fixed4.mean} for every four steps (difference \S{st.F.d}, 95\% interval \S{st.F.ci}). With the same nominal computation per step, eight search rounds every four steps reached \S{st.eq.r8} successes in a post hoc comparison with an interval for the difference that included zero; increasing candidates instead still favoured every-step replanning. At the prespecified settings, no rule showed a clear success advantage over its rate-matched control. Some rules did better at lower thresholds. Across the tested settings, no rule showed a clear success advantage over its schedule-shuffled control. The shuffled control itself outperformed the rate-matched control in \S{s3.st.sr.nholm} of 13 settings after correction. Under milder changes to simulator damping, every-step successes fell to \S{s3.mild.fixed1.min}--\S{s3.mild.fixed1.max} of 64 while the simulator-scored planner finished \S{s3.mild.oracle.min}--\S{s3.mild.oracle.max}. The comparison separates replanning rate from the pairing of replanning times with individual tasks.
\end{abstract}

\section{Introduction}

Model-predictive control with a learned dynamics model alternates between two activities: searching over candidate action sequences with the model, and executing part of the best sequence before searching again \citep{chua2018pets,hafner2019planet,hansen2024tdmpc2}. Searching is where the computation goes. A planner that searches at every control step uses the most recent observation at all times and pays for a full search each step; a planner that commits to several actions saves computation and risks acting on a stale prediction. How often to replan is a scheduling decision.

Recent work lets that decision depend on what the model observes or expects. AdaReP replans when the state has drifted from the cached rollout, adapts the threshold online, and reports large reductions in planner queries at comparable task performance \citep{cheng2026adarep}. ContactGuard forecasts the outcome of a planned action chunk before contact and stops execution when failure looks likely \citep{zheng2026contactguard}. Neither proposes replanning at anticipated contact. We construct such a rule ourselves to test an intuition that is easy to accept for manipulation, and that we do not attribute to either work: predictions are most fragile around contact, so that may be when a controller should look again.

This paper does not propose a schedule. It asks what a timing rule has to beat before its timing can be credited, and it contributes a protocol that answers the question together with one worked application. A rule that replans more often will usually do better for that reason alone. We therefore compare each rule, across a range of thresholds, with three references: the curve of success against replans per step traced by fixed-period schedules; a control that replans at random moments at the rule's own realised rate on the same tasks; and a control that runs each task with the schedule the rule produced on another task, which matches the rule at every step of the episode and not only on average.

The application is planar pushing \citep{chi2023diffusion} with an ensemble dynamics model and a sampling planner. The study has three stages. Stage 1 was a pilot with one training seed. Stage 2 was planned in writing to address limitations of the pilot: \S{st.seeds} training seeds for state input and \S{px.seeds} for image input, fresh test tasks, a threshold sweep, rate-matched controls with 20 timing draws each, cost measured per executed step instead of per task, and arms that separate the replanning period from search effort. The Stage 2 results showed that the rate-matched control left the placement of replans within the episode unmatched. Stage 3, again planned in writing before it was run, added the schedule-shuffled control and a graded set of dynamics changes. We read four things from Stages 2 and 3.

\emph{Frequency.} Under the training dynamics, mean success rises as the fixed period shortens, and every-step replanning beats the four-step period in every seed (Section~\ref{sec:freq}).

\emph{Frequency or search effort.} A four-step schedule that receives the same nominal computation per step as more search rounds is not distinguishable from every-step replanning; when it receives more candidates instead, it stays behind. A planner with exact rollouts loses about as many tasks between period 1 and period 4 as the learned model does. In this system much of the benefit of replanning often can be had by searching more (Section~\ref{sec:effort}). The contrast on which this rests was not named in the plan.

\emph{Timing.} At the threshold fixed in advance, no rule beat its rate-matched random control on successes; at lower thresholds and on best coverage several did (Section~\ref{sec:timing}). The schedule-shuffled control removes most of that advantage. No rule is distinguishably ahead of it on successes at any threshold, and it is ahead of the rate-matched control in most arms. The rules' advantage over random timing is thus largely reproduced by a schedule borrowed from another task, which replans early in the episode as the rule's own schedule does (Section~\ref{sec:shuffle}).

\emph{Changed dynamics.} Small changes of the damping parameter degrade the learned controller step by step while a planner with exact rollouts still finishes \S{s3.mild.oracle.min} to \S{s3.mild.oracle.max} of 64 tasks (\S{or.fixed1.0} in training), so over that range most of the loss is the model's. At larger values the planner fails with exact rollouts too (Section~\ref{sec:shift}).

Section~\ref{sec:limits} discusses the limits of these comparisons.

\section{Related Work}
\label{sec:related}

\paragraph{Planning with learned dynamics.} PETS combines a bootstrap ensemble of probabilistic dynamics models with sampling-based planning \citep{chua2018pets}. PlaNet plans in a learned latent space from pixels \citep{hafner2019planet}, and TD-MPC2 scales latent model-predictive control across many tasks \citep{hansen2024tdmpc2}. DINO-WM plans at test time with a world model built on pretrained visual features \citep{zhou2024dinowm}, and LeWorldModel trains a compact joint-embedding world model end to end and plans with it \citep{maes2026lewm}; both evaluate on PushT. Neither studies how often to replan. DINO-WM executes the first few actions of each plan and reports one comparison of open-loop execution with replanning; LeWorldModel executes its whole five-step plan before planning again. Our model is far smaller than these and predicts an explicit object pose. The scheduling question arises for any planner that caches a plan. What carries over to a latent model by construction is the accounting and the references: counting model predictions per executed step, the fixed-period curve, and random controls built from a rule's own replanning record, none of which reads the state. The execution-error and disagreement rules need only a predicted and an observed representation and an ensemble. The other two rules (current geometry, and the features behind the near-contact label) rely on an explicit pose and a known outline. Whether any of our readings carries over to latent models has not been tested. Keeping the remainder of the previous plan among the candidates, as we do, follows common practice in sampling-based planning \citep{pinneri2020icem}, and retraining a dynamics model on data gathered by its own controller has been analysed for model-based reinforcement learning \citep{ross2012agnostic}.

\paragraph{Evaluating world models for control.} Model error compounds over a rollout, and the horizon over which a model can be trusted is a central concern of model-based reinforcement learning \citep{janner2019mbpo}. Two recent studies evaluate world models by their response to actions: one compares the local action sensitivity of a pretrained world model with that of the simulator \citep{vasilev2026geometry}, and one ranks candidate pushes on PushT, reports decision regret (the loss from choosing the model's preferred action instead of the best one), and finds that object-centric coordinates reduce it \citep{samad2026frame}. We follow the second in expressing our model in the object frame. Neither study runs a control loop; we judge the model by closed-loop task outcomes. For planar pushing itself, Hogan and Rodriguez describe the pusher--slider as a hybrid system: at a single point contact the pusher either sticks to the slider or slides along it in one of two directions, the dynamics are discontinuous at the transitions between these modes, and together with the limited forces that contact can transmit this is what they identify as complicating feedback control \citep{hogan2016pusher}. Data-driven models of pushing have also been built \citep{bauza2017pushing}. Our near-contact label is a distance threshold; it does not distinguish sticking from sliding.

\paragraph{When to replan.} Deciding when to recompute a control input is the subject of event-triggered and self-triggered control \citep{astrom2002lebesgue,heemels2012event}. In that vocabulary an event-triggered rule recomputes when a measured quantity crosses a threshold, and a self-triggered rule decides the next recomputation time from the model at the moment of planning. Two of our rules are event-triggered in this sense: execution error and current geometry read the current observation. The other two, ensemble disagreement and predicted near-contact change, are computed from the cached plan alone and are self-triggered. That literature usually asks for stability or performance guarantees under a bound on the model error, which we do not have for a learned model; we ask the empirical question only.

For learned models, AdaReP is the closest prior work \citep{cheng2026adarep}. It replans when the observed state has drifted from the cached rollout by more than a threshold that tightens online with the recent drift and with a local sensitivity estimate, and it counts computation as the number of world-model queries the planner issues, which is the same kind of unit we use. It compares against replanning every step, replanning every fixed number of steps, and a fixed-threshold version of its own rule. The comparison is made at matched task performance: each method is tuned to stay within a small margin of every-step performance and the methods are then ranked by queries. A sweep over fixed periods appears in its appendix for one benchmark. Its real-robot experiments include planar pushing of a T-shaped block. Our execution-error rule uses the same kind of signal with a fixed threshold; it corresponds to that fixed-threshold baseline and is not a reproduction of AdaReP. AdaReP's sensitivity term lowers the threshold, and so brings replans forward, in phases where the monitored state moves a lot per unit of action; its example is a door near its hinge. Our predicted near-contact rule is a neighbouring idea with a different mechanism: it replans on a predicted change of contact status, read from the cached plan before the step is executed, and does not weight a measured deviation. Goh et al.\ gate a sampling planner with a fitted out-of-distribution score, falling back to an amortised policy on the remaining steps, and compare against schedules that plan on a fixed fraction of steps at matched latency; they report that state-aware gating wins in one domain and is matched or beaten by fixed-rate schedules in others \citep{goh2026thinkfast}. The trigger can also be learned. B{\o}hn et al.\ train a policy that decides at each step whether a model-predictive controller solves its problem again, jointly with the prediction horizon and a linear feedback law applied between solves, on a simulated cart-pendulum. The reward charges each solve in proportion to its horizon. The baselines are fixed recomputation intervals over a grid of horizons, and in an ablation that spends the same computation at a fixed interval the cost rises by 35\% \citep{bohn2023mpcmeta}. Honda et al.\ train a deep Q-network that decides at each step whether the global planner of a simulated navigation robot replans, from range-scan points, the current path and the past trajectory. They compare it with time-, distance- and stuck-based rules at one hand-tuned setting each and report the number of replans without matching it \citep{honda2023replan}. Neither includes a control that replans at random moments. ContactGuard forecasts the outcome of a planned action chunk shortly before a commanded grasp closure and aborts execution when failure looks likely; it does not replan. Its baselines are other failure detectors and simpler probes of the same latent model, and it reports computation as the latency of one forecasting pass. Nothing is being scheduled, so it has no fixed-interval or random-timing baseline \citep{zheng2026contactguard}. For policies that emit action chunks, the number of actions executed before the next query plays the role of our period: it is a tuned setting in action chunking \citep{zhao2023act} and in Diffusion Policy \citep{chi2023diffusion}, and test-time resampling has been proposed to make chunked execution more reactive \citep{liu2024bid}.

This comparison draws on the full texts of nine works on replanning and on evaluating world models: AdaReP, ContactGuard, Goh et al., B{\o}hn et al., Honda et al., DINO-WM, LeWorldModel and the two evaluation studies. The selection was targeted and is not a systematic search. Fixed-period baselines appear in AdaReP, in Goh et al., in B{\o}hn et al.\ and, at a single period, in Honda et al. B{\o}hn et al.\ come closest to a matched comparison: one ablation replaces their learned trigger by a fixed interval that spends the same computation, at one operating point on one simulated system, and the fixed interval does worse. That is prior evidence, on another system and measure, that timing can matter at equal computation. The comparison found no random-replanning control among these nine works and no comparison of a timing rule over a range of thresholds with a fixed-period curve at matched computation. Those are the comparisons this paper adds, on one simulated task. Nothing here contradicts the savings AdaReP reports, which concern a different planner, model and rule, compared at matched performance.

\paragraph{Changed dynamics.} Online adaptation of a dynamics model has been demonstrated with meta-learning \citep{nagabandi2019adapt}, and policies conditioned on parameters identified online have been trained for unknown dynamics \citep{yu2017uposi}. We implement neither; Section~\ref{sec:shift} reports what happens without them. Replanning also takes time on a real system, and the delay between measurement and applied input has to be accounted for in the controller \citep{findeisen2004delay}; our simulation applies each plan in the step in which it is computed.

\section{Setup}
\label{sec:setup}

\subsection{Task and the damping parameter}

We use the PushT environment of gym-pusht \citep{gympusht} in a short-range variant. A circular pusher of radius 15~px moves a T-shaped block on a $512\times512$~px table. The action is an absolute target position for the pusher, clipped to $[16,496]$~px; the pusher is a kinematic body driven by a position controller at 10~Hz with ten simulation substeps of 0.01~s per control step. Success is the environment's own criterion: the block covers more than 95\% of the goal region.

Our variant departs from the standard benchmark in three ways. The goal is not the environment's fixed pose: for each task it is the block pose that a scripted, state-reading controller reaches after 14 steps from the same start, so every goal is reachable within a short episode. Starts are confined to a narrow range, with the block near the centre of the table at a random angle within $\pm0.5$~rad and the pusher just below the stem of the T. Episodes last 35 control steps. Success rates here are therefore not comparable with published PushT results; only comparisons between schedules on the same tasks are meaningful. The goal pose is given to the controller as numbers in both input modes.

The environment exposes a damping parameter $d\in[0,1]$ that the physics engine applies to rigid-body velocity at every substep of length $\Delta t$:
\begin{equation}
v \leftarrow v\, d^{\,\Delta t}.
\end{equation}
The parameter is the fraction of velocity retained per second, so we call it velocity retention: larger values mean less dissipation and a more slippery table, and a free body's speed decays with time constant $-1/\ln d$. With $d=0$, the default, velocity is removed at every substep and the block moves only while it is pushed. This is the quasi-static regime usually assumed for planar pushing. We measured how far the block keeps travelling in the ten control steps after the pusher is told to hold its position: \S{coast.0}~px at $d=0$ (the pusher's own residual motion), and \S{coast.0.02}, \S{coast.0.05}, \S{coast.0.1}, \S{coast.0.25} and \S{coast.0.85}~px at $d=0.02$, $0.05$, $0.10$, $0.25$ and $0.85$, whose time constants are 0.26, 0.33, 0.43, 0.72 and 6.2~s. The pusher is kinematic and is not affected by $d$. We train only at $d=0$ and treat every other value as unseen dynamics.

\subsection{Dynamics model and data}

The state is the six-number pose $q=(p_x,p_y,b_x,b_y,\sin\theta,\cos\theta)$, with pusher and block positions divided by 512. Each ensemble member takes the previous pose, the current pose, the candidate action and the previously commanded action. These 16 numbers are mapped to 21 features: the current pose and, in the block frame, the relative position of the pusher, the offset of the command, the displacement over the last step, the change of angle, and the signed distances from the pusher and from the commanded target to the outline of the T, computed analytically from the known shape. A three-layer perceptron with 256 hidden units predicts the displacement of the pusher and of the block in the block frame and an increment of the block angle; these are scaled, rotated back to table coordinates and added to the current pose. A second network with two hidden layers of 96 units and separate parameters predicts the probability that the pusher will be in near-contact at the next step, where near-contact means that the signed distance from the pusher centre to the outline is at most 15.5~px. With a pusher of radius 15~px this marks touching or a gap below half a pixel. The label is geometric: it is not a contact force, and it does not distinguish making or breaking contact from sustained pushing. Each member has \R{st.params} parameters.

The ensemble has three members, each trained on a bootstrap resample as in PETS \citep{chua2018pets}. Unlike the probabilistic networks of PETS, each member here is deterministic and outputs a point prediction. Deep ensembles, whose members differ by random initialisation, use the spread between members as an uncertainty signal \citep{lakshminarayanan2017ensembles}, and we use disagreement between our members in the same way. The pose loss is the squared error of the predicted next pose, with positions divided by 0.02, the sine and cosine divided by 0.05, and the angular term weighted by 0.1; the near-contact network is trained with binary cross-entropy. Each network keeps the epoch with its own best validation loss. Appendix~\ref{app:hyper} lists the training settings.

In image mode a convolutional network with \R{px.visparams} parameters estimates the same pose from the current $64\times64$ RGB frame, after the goal overlay is painted out so that the goal is never read from the image. The dynamics ensemble is then trained on estimated poses as inputs and true next poses as targets.

Data come from the simulator at $d=0$ only. We first collect \R{data.explore} exploration episodes of 32 steps with four scripted behaviours, then run three rounds in which the current controller collects 1,200 further episodes and the model is retrained on everything, in the manner of dataset aggregation \citep{ross2011dagger}. Collection uses task seeds disjoint from calibration and test, and adds Gaussian noise with a standard deviation of 10~px to the action on 30\% of steps. Each final set has \R{data.episodes} episodes and \R{data.transitions} transitions, split by episode into training, validation and test in proportions 7:2:1.

For the pilot model (training seed 23) the one-step position error on \R{st.testtrans} held-out transitions is \R{st.rmse}~px, as a root mean square over the four position coordinates. Its near-contact network reaches an area under the receiver operating characteristic curve (AUROC) of \R{st.auroc} and a Brier score of \R{st.brier}, where the Brier score is the mean squared difference between the predicted probability and the 0/1 label; a constant prediction at the base rate scores \R{st.base}. The image front-end has a position error of \R{px.vis.rmse}~px and a mean angle error of \R{px.vis.ang} degrees. Section~\ref{sec:where} breaks the dynamics error down by type of transition for all ten state-input models.

\subsection{Planner and the unit of computation}

At each replan the controller scores $K=32$ candidate sequences of $H=6$ free actions. Up to eight candidates are structured: a straight push along the block-to-goal direction with a stride equal to half the remaining distance, limited to between 3 and 28~px; holding still; moving to a standing point 95~px behind the block's centre; four targets that orbit the block's centre, at $\pm35^\circ$ and $\pm70^\circ$ from the pusher's current bearing and at the pusher's current distance or 95~px if that is larger; and the remainder of the cached plan. Of the rest, three quarters are Gaussian perturbations of the straight push or of the cached plan at scales from 1.4 to 23~px, and one quarter are constant reach targets within 133~px of the block. After the first round of scoring, the five best candidates define a mean and spread from which a second round is sampled, one step of the cross-entropy method \citep{deboer2005cem}. Every candidate is rolled out for the six free actions followed by six further steps that repeat the last target, so that motion after release is predicted. The cost of a candidate is the pose cost at the last step plus 0.2 times its mean over the rollout, where the pose cost is the squared position error of the block plus $0.1\,(1-\cos\Delta\theta)$, and a tie-breaking term: 0.3 times the current pose cost times the squared distance between the standing point and the pusher position predicted after the last free action. The ensemble mean is used; no disagreement penalty is applied.

We count computation in \emph{prediction units}: one unit is one ensemble member predicting one candidate for one step. A full plan costs $3$ members $\times\,32$ candidates $\times\,12$ rollout steps $\times\,2$ search rounds $=2{,}304$ units. Three things lie outside the unit. The controller reads an observation at every step under every schedule; in image mode this runs the front-end once per step whether or not the controller replans. Monitoring for the timing rules reuses the cached plan and costs no units. And the unit is not a statement about time: on one thread of the laptop processor used here (Apple M5 Max) a full plan takes about \S{time.state.plan}~ms, well inside the 100~ms control period, and a front-end call about \S{time.pixels.obs}~ms. In this system every schedule could therefore run in real time, each replan has the same peak cost, and the schedules differ in average load. The simulation applies a plan in the step in which it is computed, with no delay.

\subsection{Schedules}
\label{sec:schedules}

The fixed-period schedules replan every 1, 2, 3, 4 or 6 steps. The four timing rules replan when a score exceeds a threshold, and in any case after six steps, the number of free actions in a plan. A step is \emph{eligible} for a timing rule when a cached plan exists and fewer than six of its actions have been executed.
\begin{itemize}
\item \emph{Execution error} (measurement-based): the root-mean-square difference between the observed positions and those the cached plan predicted for the current step.
\item \emph{Current geometry} (measurement-based): whether the pusher is within 17.5~px of the outline according to the current pose estimate, compared with its status when the plan was made. It uses no model and has no threshold.
\item \emph{Ensemble disagreement} (prediction-based): the cached standard deviation across members for the step about to be executed.
\item \emph{Predicted near-contact change} (prediction-based): the range of the cached near-contact probabilities over a window of up to three steps around the step about to be executed.
\end{itemize}
The two measurement-based rules need the current observation; the two prediction-based rules are fixed when the plan is made. The \emph{random-timing control} replans at each eligible step with a fixed probability $p$ and, like the rules, after six steps at the latest.

Thresholds are calibrated in closed loop for each model. We run the every-six-steps schedule on 48 calibration tasks, record the three scores at every step, and take percentiles of each. Once a rule is active, each trigger replaces the plan and lowers the scores, so realised trigger rates differ from the nominal ones.

\subsection{Three stages}
\label{sec:stages}

The archived reference experiments cover three stages.

\emph{Stage 1} is a pilot with one training seed and 64 test tasks. Each rule was run at one threshold, the 75th percentile, and the random control's probability was set to the rate at which the near-contact rule fired on the calibration tasks. Five paired comparisons were fixed in the analysis code beforehand. Stage 1 also ran every schedule under a cap of twelve full plans and under two unseen damping values. The Stage 1 design had four limitations: the random control was not matched to the rule on the test tasks; computation per task depended on the outcome, because episodes ended at success; each rule was tested at one threshold; and replanning more often also meant searching more. Stage 1 is reported in Section~\ref{sec:stage1} as exploratory.

\emph{Stage 2} was designed to answer those points, and its plan was written down before any Stage 2 run (Appendix~\ref{app:plan}). It uses fresh test tasks, \S{st.seeds} training seeds for state input and \S{px.seeds} for image input, and the following arms for every model.
\begin{itemize}
\item Fixed periods 1, 2, 3, 4 and 6.
\item Each of the three thresholded rules at the 25th, 50th, 75th and 90th percentile of its calibration scores, and the geometry rule: thirteen rule arms.
\item A rate-matched random-timing control for each rule arm (below).
\item Equal-budget arms that separate the period from the search effort per step: periods 1, 2 and 4, each at 576, 1,152 and 2,304 prediction units per step, obtained by changing the number of candidates from 8 to 128; and period 4 at 2,304 units per step with more search rounds instead of more candidates.
\end{itemize}
Seeds share the exploration set and differ in ensemble initialisation, bootstrap resampling and their own three rounds of on-policy data. Episodes run for the full 35 steps; success is coverage above 0.95 at any step, and the trajectory up to the first success is identical to that of an episode that stops there, so Stage 1 and Stage 2 outcomes are defined alike.

\paragraph{Cost axis.} The primary measure of cost is \emph{replans per executed step}, counted up to the first success or step 35, and the same number multiplied by the units per plan. For a fixed period and for the random control it is set by the schedule and not by the outcome; for a timing rule it can still depend on when the episode ends, because a rule need not replan at an even rate. We also report units over the full 35-step episode. Mean units per task up to the first success, the axis of Stage 1, is given for reference only.

\paragraph{Rate-matched random control.} For a constant probability $p$ and a six-step limit, the long-run number of replans per step is $p/(1-(1-p)^6)$. For each model and each rule arm we set $p$ so that this equals the rule's realised replans per executed step on the test tasks, run four timing draws per task, read only their replans per executed step, and adjust $p$ once by the difference. The control reported is 20 further independent timing draws per task at the adjusted $p$. Outcomes of the tuning draws are never read. A control whose realised rate differs from the rule's by more than 0.015 replans per step is flagged as not matched and is not tuned again.

\emph{Stage 3} addressed a gap in the rate-matched control. The Stage 2 results showed that the control equalises replans per executed step over the episode but not when in the episode the replans fall: in steps 0 to 9, where most successes are decided, the rules replanned more often than their controls (Section~\ref{sec:shuffle}). Stage 3 was planned in writing before any of its episodes was run (Appendix~\ref{app:plan}). It adds a second control for every rule arm and model, four milder values of $d$ (Section~\ref{sec:shift}), and a measurement of the execution-error score while the block is at rest (Appendix~\ref{app:hyper}).

\paragraph{Schedule-shuffled control.} Stage 2 recorded, for every model, rule arm and test task, the rule's schedule as a string of 35 bits: replan or reuse at each step of the full episode. For task $i$ and draw $k$ the shuffled control runs task $i$ with the schedule that the same rule and the same model produced on another task $\pi_k(i)$, where $\pi_k$ is a random permutation of the 64 tasks without fixed points. The controller replans exactly where the borrowed schedule says so. Planner, plan size and planner random numbers are those of the rule's own episode on task $i$, and there are 20 draws per task. Within a draw every schedule is used once, so over the tasks the control makes as many replans as the rule at every step of the episode, with the same runs and gaps. What it removes is the link between a schedule and the state of the episode it is applied to. Replaying a rule's schedule on its own task reproduces the rule's episode exactly, which the code checks on two tasks per rule arm for every model. Because rule minus rate-matched control equals rule minus shuffled control plus shuffled control minus rate-matched control, the two parts separate what a rule gains by responding to the state from what it gains through the shape of its schedule.

\subsection{Evaluation and inference}

Development tasks (16) were used to design the planner and calibration tasks (48) to set thresholds. Stage 1 used 64 test tasks; Stage 2 uses 64 different test tasks, the same for every model, so seeds and tasks are crossed. Within a model all arms see the same tasks with the same planner randomness.

The unit of analysis in Stage 2 is a (seed, task) cell. We report the mean number of successes out of 64 over seeds, with the range over seeds, and \emph{best coverage}, the highest coverage reached in the episode, as a continuous outcome. Intervals are 95\% percentile intervals from a crossed bootstrap that resamples seeds and tasks with replacement (10,000 resamples), so they include variation between seeds and between tasks; $p$-values are two-sided and come from the same resamples. For a comparison with a control, the control's value in a cell is its mean over its 20 draws.

The primary family has five comparisons on state input, corrected by Holm's method: every step against every four steps, and each of the four rules against its rate-matched control, the thresholded rules at the 75th percentile. All other Stage 2 comparisons are secondary and uncorrected: the remaining thresholds against their controls, every rule arm against the fixed-period curve at the same replans per executed step (linear interpolation between neighbouring periods), the equal-budget arms, and everything on image input. We call two arms not distinguishable when the interval for their difference includes zero, and we always report the interval.

The plan named best coverage as a co-primary outcome but defined the Holm family on successes only. We apply the same five comparisons and the same correction to best coverage and report the two families side by side. Appendix~\ref{app:plan} lists this and the other points at which the analysis goes beyond the written plan.

Stage 3 has its own primary family, fixed in its plan: each of the four rules at the same settings minus its shuffled control, on successes and on best coverage, eight comparisons corrected together by Holm's method. The plan also fixed how the result would be worded. A rule's timing is credited only where rule minus shuffled control excludes zero after the correction. Where the interval for rule minus shuffled control includes zero and that for shuffled minus rate-matched control excludes zero, the rule's Stage 2 advantage is attributed to the shape of its schedule. Where both include zero, the advantage is reported as not resolved. The sweep against the shuffled control and the comparison of the two controls are secondary.

\paragraph{Protocol.} The procedure can be applied to another planner in six steps. (1) Fix a unit of computation that the planner spends per replan, and report cost per executed step. Cost per task depends on the outcome when episodes end at success; count cost per executed step up to that step and say so, or run episodes of fixed length. (2) Run fixed-period schedules to trace success against cost. (3) Run the timing rule over a range of thresholds, not at one. (4) For each rule arm, run a random-timing control whose probability is tuned on the evaluation tasks, using only the realised replanning rate, until it matches the rule; use many timing draws. (5) For each rule arm, run a schedule-shuffled control: record the rule's schedule on every evaluation task and run each task with the schedule from another. (6) Report every difference with an interval that covers tasks and training seeds, on the threshold outcome and on a continuous one, and with a null say which advantages the interval excludes and which it does not. A rule's timing is credited only if it beats its shuffled control. Beating the rate-matched control shows that the rule's schedule is better than an evenly spread one at the same rate, which may be due to its shape alone. Lying above the fixed-period curve shows less still, because a schedule with irregular spacing need not lie on that curve. In our application the three references gave three different verdicts (Sections~\ref{sec:timing} and~\ref{sec:shuffle}).

\section{Results}

Unless stated otherwise, results are from the archived Stage 2 reference runs at $d=0$ with state input: \S{st.seeds} models, 64 test tasks, \S{st.cells} seed-task cells.

\subsection{Frequency}
\label{sec:freq}

Table~\ref{tab:main2} gives the fixed periods and the four rules at their pre-specified threshold. The mean number of successes falls from \S{st.fixed1.mean} of 64 (range over seeds \S{st.fixed1.min} to \S{st.fixed1.max}) when the controller replans every step to \S{st.fixed6.mean} (\S{st.fixed6.min} to \S{st.fixed6.max}) when it replans every six steps. Every step against every four steps, the first comparison of the primary family, gives a difference of \S{st.F.d} tasks (interval \S{st.F.ci}, \S{st.F.hh}; Table~\ref{tab:primary}), and every-step replanning is ahead in all \S{st.f1gtf4} seeds (Table~\ref{tab:perseed}). On best coverage the difference is \S{st.Fcov.d} (interval \S{st.Fcov.ci}). The reading does not depend on the success threshold: with thresholds of 0.90 and 0.97 the difference is \S{st.Fs90.d} and \S{st.Fs97.d} tasks (intervals \S{st.Fs90.ci} and \S{st.Fs97.ci}; Table~\ref{tab:sens}). Success and best coverage both record a first passage or a maximum, which more frequent replanning could reach by chance without holding the block there. Coverage at the last step of the episode is neither: it is \S{s3.st.final.fixed1} against \S{s3.st.final.fixed4} (difference \S{s3.st.final.F.d}, interval \S{s3.st.final.F.ci}), and \S{s3.st.final95.fixed1} against \S{s3.st.final95.fixed4} episodes end above 0.95, a comparison added during the Stage 3 analysis. Between neighbouring periods the differences are \S{st.pair.fixed1.fixed2.d} (period 1 against 2; interval \S{st.pair.fixed1.fixed2.ci}), \S{st.pair.fixed2.fixed3.d} (2 against 3; \S{st.pair.fixed2.fixed3.ci}), \S{st.pair.fixed3.fixed4.d} (3 against 4; \S{st.pair.fixed3.fixed4.ci}) and \S{st.pair.fixed4.fixed6.d} (4 against 6; \S{st.pair.fixed4.fixed6.ci}).

The cost of this gain depends on the axis. Replanning every step makes \S{st.ratio.rate} times as many replans per executed step as replanning every four steps, and \S{st.ratio.u35} times as many over a full 35-step episode. On the Stage 1 axis, units up to the first success, the ratio is only \S{st.ratio.ustop}, because the every-step schedule succeeds more often and its successful episodes stop counting early. Most successes arrive early in the episode: of the \S{st.fixed1.mean} successes of the every-step schedule, \S{cum.fixed1.10} have occurred by step 10 (Figure~\ref{fig:when}).

\begin{table}[t]
\centering
\caption{State input, $d=0$: mean successes of 64 over \S{st.seeds} training seeds, with the range over seeds. Each timing rule is shown at its 75th-percentile threshold above its rate-matched random-timing control (20 timing draws per task). Units per step are prediction units per executed step.}
\label{tab:main2}
\small
\begin{tabular}{lrrrr}
\toprule
Schedule & Replans per step & Units per step & Successes of 64 & Best coverage \\
\midrule
Every step & 1.00 & 2304 & 33.7 (28--38) & 0.936 \\
Every 2 steps & 0.52 & 1190 & 29.4 (23--36) & 0.928 \\
Every 3 steps & 0.35 & 799 & 21.4 (17--27) & 0.909 \\
Every 4 steps & 0.26 & 602 & 16.8 (15--19) & 0.903 \\
Every 6 steps & 0.17 & 402 & 12.9 (10--15) & 0.886 \\
\midrule
Execution error & 0.23 & 537 & 18.0 (15--23) & 0.907 \\
\quad rate-matched random timing & 0.23 & 527 & 16.0 (13--20) & 0.895 \\
Ensemble disagreement & 0.23 & 541 & 17.7 (11--23) & 0.908 \\
\quad rate-matched random timing & 0.23 & 532 & 16.1 (12--19) & 0.895 \\
Predicted near-contact & 0.34 & 781 & 19.7 (16--23) & 0.913 \\
\quad rate-matched random timing & 0.34 & 775 & 21.4 (19--26) & 0.909 \\
Current geometry & 0.26 & 588 & 20.1 (18--23) & 0.913 \\
\quad rate-matched random timing & 0.25 & 577 & 17.4 (14--20) & 0.898 \\
\bottomrule
\end{tabular}

\end{table}

\begin{table}[t]
\centering
\caption{The five planned comparisons of Stage 2, state input; the control is the rate-matched one. Intervals are 95\% intervals from a crossed bootstrap over seeds and tasks. The Holm family on successes was fixed in advance; the same family on best coverage was added in the analysis.}
\label{tab:primary}
\small
\begin{tabular}{lrrrr}
\toprule
 & \multicolumn{2}{c}{Successes of 64} & \multicolumn{2}{c}{Best coverage} \\
Comparison & Difference (95\% interval) & Holm $p$ & Difference (95\% interval) & Holm $p$ \\
\midrule
Every step $-$ every 4 steps & +16.9 (+11.6 to +22.2) & <0.001 & +0.033 (+0.023 to +0.044) & <0.001 \\
Execution error $-$ control & +2.0 (\ensuremath{-}1.1 to +5.5) & 0.746 & +0.012 (+0.005 to +0.019) & <0.001 \\
Disagreement $-$ control & +1.6 (\ensuremath{-}1.3 to +4.5) & 0.746 & +0.013 (+0.005 to +0.021) & <0.001 \\
Near-contact $-$ control & \ensuremath{-}1.6 (\ensuremath{-}5.7 to +2.3) & 0.746 & +0.004 (\ensuremath{-}0.004 to +0.013) & 0.371 \\
Geometry $-$ control & +2.7 (\ensuremath{-}1.3 to +6.8) & 0.746 & +0.015 (+0.008 to +0.023) & <0.001 \\
\bottomrule
\end{tabular}

\end{table}

\subsection{Frequency or search effort}
\label{sec:effort}

\begin{figure}[t]
\centering
\includegraphics[width=0.92\linewidth]{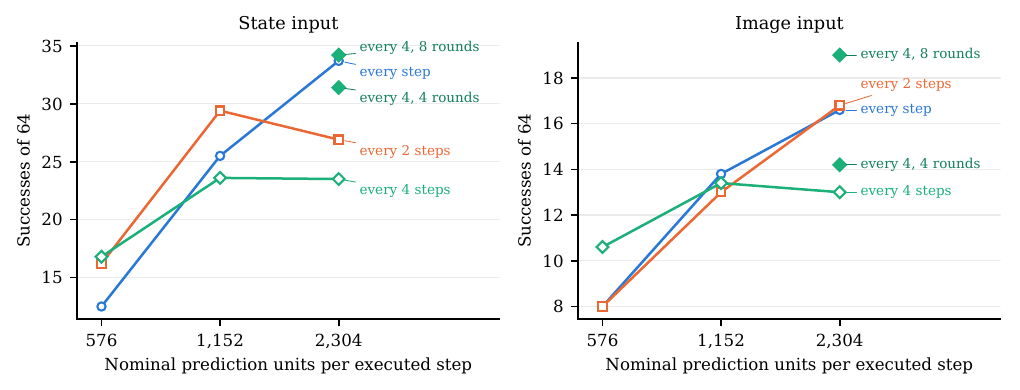}
\caption{Successes against nominal prediction units per executed step for periods 1, 2 and 4. Open markers change the plan size through the number of candidates; filled markers give the four-step period the same 2,304 units per step through more search rounds.}
\label{fig:equal}
\end{figure}

With a plan of fixed size, units per step are inversely proportional to the period, so a schedule that replans more often also searches more. The equal-budget arms hold units per executed step constant while the period changes (Table~\ref{tab:equal}, Figure~\ref{fig:equal}).

When the extra computation is given to the slower schedules as more candidates, the picture depends on the budget. At 576 units per step, replanning every step with 8 candidates succeeds on \S{st.eq.576.1} tasks and every four steps with 32 candidates on \S{st.eq.576.4} (period 1 minus period 4: \S{st.pair.period.576.d}, interval \S{st.pair.period.576.ci}). At 1,152 units the three periods give \S{st.eq.1152.1}, \S{st.eq.1152.2} and \S{st.eq.1152.4} (\S{st.pair.period.1152.d}, \S{st.pair.period.1152.ci}). Only at 2,304 units does every-step replanning lead, \S{st.eq.2304.1} against \S{st.eq.2304.4} (\S{st.pair.period.2304.d}, \S{st.pair.period.2304.ci}). More candidates do not clearly help the four-step schedule: going from 32 to 64 changes its successes by \S{st.pair.cand.k64.fixed4.d} tasks (\S{st.pair.cand.k64.fixed4.ci}) and going from 64 to 128 by \S{st.pair.cand.k128.k64.d} (\S{st.pair.cand.k128.k64.ci}). The planned contrasts between 2,304 and 576 units per step at a fixed period are \S{st.pair.budget.p1.d} tasks at period 1 (\S{st.pair.budget.p1.ci}) and \S{st.pair.budget.p4.d} at period 4 (\S{st.pair.budget.p4.ci}).

When the same computation is given as more search rounds, the lead is no longer distinguishable. Replanning every four steps with 32 candidates and eight rounds, which nominally costs 2,304 units per step like every-step replanning with two rounds, succeeds on \S{st.eq.r8} tasks against \S{st.eq.2304.1} (difference \S{st.pair.rounds.r8.fixed1.d}, interval \S{st.pair.rounds.r8.fixed1.ci}); with 64 candidates and four rounds it succeeds on \S{st.eq.r4} (\S{st.pair.rounds.r4.fixed1.d}, \S{st.pair.rounds.r4.fixed1.ci}). At the same period and the same units, eight rounds beat 128 candidates by \S{st.pair.rounds.r8.k128.d} tasks (\S{st.pair.rounds.r8.k128.ci}). Raising the four-step schedule from two to eight rounds adds \S{st.pair.rounds.r8.fixed4.d} tasks (\S{st.pair.rounds.r8.fixed4.ci}), about as much as moving from the four-step period to every step. The budgets are nominal: a four-step schedule plans at steps 0, 4, 8 and so on of an episode that may end in between, so its realised cost per executed step is a few per cent higher, \S{st.fixed4K32i8.ups} units for the eight-round arm against \S{st.fixed1.ups} for every-step replanning.

The simulator-scored planner points the same way. It keeps the proposals, the cost and the horizon and replaces the learned model by the simulator when scoring candidates; it reads the simulator state and is not a deployable controller. With exact rollouts the predictions of the cached plan come from the simulator itself, so a replan has little or no prediction error to correct. It still succeeds on \S{or.fixed1.0}, \S{or.fixed2.0}, \S{or.fixed3.0}, \S{or.fixed4.0} and \S{or.fixed6.0} of 64 tasks at periods 1, 2, 3, 4 and 6 (period 1 minus period 4: \S{or.pair.fixed1.fixed4.d}, interval over tasks \S{or.pair.fixed1.fixed4.ci}). Between periods 1 and 4 that difference is as large as the one for the learned model; between periods 1 and 6 it is larger (\S{or.pair.fixed1.fixed6.d} tasks). What a replan adds here is another search and a horizon that moves forward with the block.

We read these results as follows. In this system much of the benefit of replanning often can be had by searching more, and the benefit cannot be taken as evidence that frequent replanning corrects model error. Three limits apply. The comparison with eight rounds is a difference whose interval includes zero, not a demonstrated equality, and the planned comparison, in which the four-step schedule receives more candidates, leaves every-step replanning ahead. We did not run every-step replanning with eight rounds, so we do not know whether frequency adds anything on top of rounds. And the contrasts that involve rounds, candidate counts and the 1,152-unit budget were not named in the written plan, although their arms were.

The simulator-scored planner also measures what the learned model costs: with every-step replanning it succeeds on \S{or.fixed1.0} of 64 tasks against \S{st.fixed1.mean} for the learned model.

\begin{table}[t]
\centering
\caption{Equal nominal computation per executed step, state input: mean successes of 64 over \S{st.seeds} seeds. Budgets are nominal; realised units per executed step are a few per cent higher for the slower schedules (Section~\ref{sec:effort}). In the first three rows the plan size is changed through the number of candidates (8 to 128) with two search rounds.}
\label{tab:equal}
\small
\begin{tabular}{lrrr}
\toprule
Units per step & Every step & Every 2 steps & Every 4 steps \\
\midrule
576 & 12.5 & 16.2 & 16.8 \\
1,152 & 25.5 & 29.4 & 23.6 \\
2,304 & 33.7 & 26.9 & 23.5 \\
\midrule
2,304, as 64 candidates $\times$ 4 rounds & & & 31.4 \\
2,304, as 32 candidates $\times$ 8 rounds & & & 34.2 \\
\bottomrule
\end{tabular}

\end{table}

\subsection{Timing rules against rate-matched random timing}
\label{sec:timing}

\paragraph{Matching.} For all \S{st.seeds} models and all thirteen rule arms the control's realised replans per executed step fell within 0.015 of the rule's; no arm is flagged (Table~\ref{tab:sweep}, third column). The match is on that pooled rate only. Over full 35-step episodes, which continue past the first success, the rules replan as often as their controls or less often, by up to 0.10 replans per step (rule minus control between \S{s3.st.rate35.random.min} and \S{s3.st.rate35.random.max}); and within the episode the rules and their controls replan at different times, which Section~\ref{sec:shuffle} takes up.

\paragraph{Planned comparisons.} At the 75th-percentile threshold the differences in successes between rule and rate-matched control are \S{st.T-error.d} for execution error (interval \S{st.T-error.ci}), \S{st.T-disagreement.d} for ensemble disagreement (\S{st.T-disagreement.ci}), \S{st.T-contact.d} for predicted near-contact (\S{st.T-contact.ci}) and \S{st.T-geom.d} for current geometry (\S{st.T-geom.ci}); every Holm-corrected $p$ is \S{st.T-error.holm} (Table~\ref{tab:primary}). No rule is distinguishable from random timing at the same rate. The intervals are much narrower than in the pilot: they exclude an advantage above \S{st.prim.maxhi} tasks and a disadvantage beyond \S{st.prim.minlo} tasks in 64. For scale, neighbouring fixed periods differ by \S{st.pair.fixed4.fixed6.d} to \S{st.pair.fixed2.fixed3.d} tasks, so an advantage as large as one step along the fixed-period axis is not excluded.

On best coverage the same comparisons give \S{st.T-errorcov.d} (\S{st.T-errorcov.ci}), \S{st.T-disagreementcov.d} (\S{st.T-disagreementcov.ci}), \S{st.T-contactcov.d} (\S{st.T-contactcov.ci}) and \S{st.T-geomcov.d} (\S{st.T-geomcov.ci}). Execution error, disagreement and geometry are ahead of their controls; the near-contact rule is not. These three differences survive Holm's correction within the coverage family and across all ten comparisons on both outcomes (largest corrected $p$ \S{st.holm10.max}). The coverage family was defined during the analysis, after the data had been seen, and by the plan's own decision rule, which refers to successes, the timing result is a null. The differences are small. The gain in coverage from replanning every step instead of every four steps is \S{st.Fcov.d}, at \S{st.ratio.rate} times the replans; the rules gain a third to a half of that at the same pooled rate. The two outcomes do not contradict each other: the point estimates for successes are positive for the same three rules, and a shift of about 0.01 in coverage moves few episodes across the 0.95 threshold.

\begin{figure}[t]
\centering
\includegraphics[width=\linewidth]{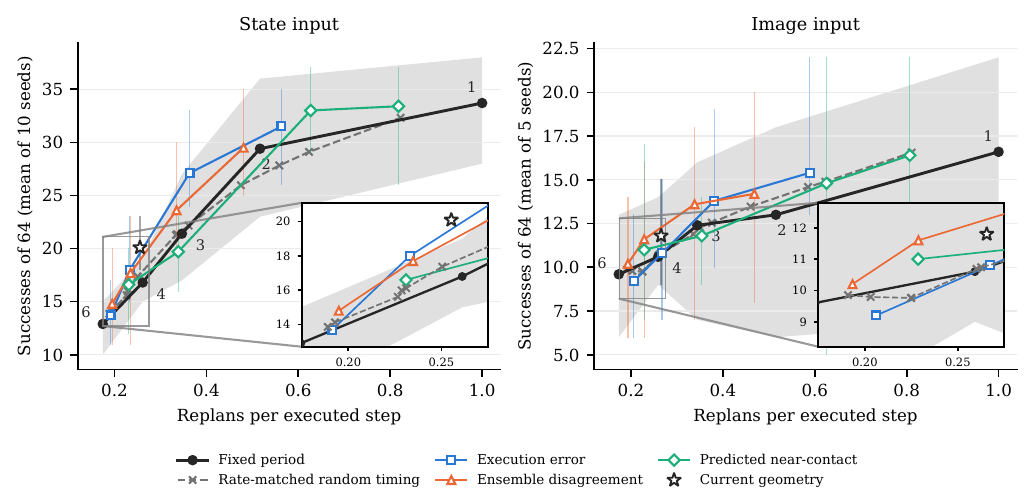}
\caption{Successes against replans per executed step. Fixed periods (numbers give the period; the band is the range over seeds) and the rate-matched random-timing controls are two of the references. Each thresholded rule is traced over four thresholds; thin vertical lines give the range over seeds for the rule arms. The inset enlarges the region that holds the 75th- and 90th-percentile thresholds.}
\label{fig:curve2}
\end{figure}

\paragraph{Threshold sweep.} Table~\ref{tab:sweep} and Figure~\ref{fig:curve2} show every rule arm. These comparisons are secondary and were planned without correction. On successes, \S{st.sw.ndpos} of the 13 differences from the rate-matched control are positive and \S{st.sw.npos} intervals exclude zero, all at the 25th or 50th percentile: execution error (\S{st.sw.errorq25.d} and \S{st.sw.errorq50.d}), disagreement at the 25th percentile (\S{st.sw.disagreementq25.d}) and near-contact at the 50th (\S{st.sw.contactq50.d}). No arm is distinguishably below its control; two point estimates are negative. If Holm's correction is applied across the thirteen, which the plan did not ask for, \S{st.sw.nholm} remain: execution error and near-contact at the 50th percentile. Against the fixed-period curve, \S{st.swc.nsig} arm is distinguishable, execution error at the 50th percentile (\S{st.swc.errorq50.d}, \S{st.swc.errorq50.ci}): it reaches \S{st.errorq50.mean} successes at \S{st.errorq50.rate} replans per step, where the two-step period needs \S{st.fixed2.rate} for \S{st.fixed2.mean}. On best coverage \S{st.swcov.npos} of the 13 arms are ahead of their controls and none is behind (Table~\ref{tab:sweepcov2}). The near-contact rule at the 50th percentile is among them: it exceeds its control by \S{st.sw.contactq50.d} tasks and by \S{st.swcov.contactq50.d} in coverage.

The column ``Signalled'' in Table~\ref{tab:sweep} helps to read the pattern. A rule replans either because its score crossed the threshold or because six steps have passed. At the 75th percentile only \S{st.errorq75.sig}\% of the execution-error rule's replans are signalled; the rest are forced by the six-step limit, and the schedule replans \S{st.errorq75.rate} times per step, not far above the every-six-steps schedule's \S{st.fixed6.rate}. At the 50th percentile \S{st.errorq50.sig}\% are signalled. A rule that seldom fires has little timing to credit. This description was added during the analysis; we did not test it as an explanation, and the near-contact rule does not follow it (\S{st.contactq75.sig}\% signalled at the 75th percentile, with a negative point estimate).

\paragraph{Random timing against the fixed-period curve.} None of the 13 rate-matched controls is distinguishable from the fixed-period curve at its own rate (differences from \S{st.swk.dmin} to \S{st.swk.dmax} tasks; a comparison added during the analysis). The intervals are wide, for example \S{st.swk.errorq25.ci} tasks for the control of the execution-error rule at the 25th percentile, so this is an absence of evidence and not a demonstration that the two references are equivalent (Table~\ref{tab:ctrlcurve} gives every interval). They can give different verdicts. Of the four arms that exceed their control, only one exceeds the curve, because in the other three the control sits below the curve (by \S{st.swk.errorq25.d}, \S{st.swk.disagreementq25.d} and \S{st.swk.contactq50.d} tasks). The pilot's observation that its random control sat above the curve is not confirmed.

\paragraph{What these comparisons hold fixed.} Taken alone, this subsection supports a cautious positive reading: at the threshold fixed in advance a null on successes with a small advantage in coverage, and at lower thresholds an advantage in successes for execution error, disagreement and near-contact. All of it is relative to a control that replans with the same probability at every eligible step of the episode. The rules do not replan that way.

\begin{table}[t]
\centering
\caption{Threshold sweep, state input, successes of 64. ``Pct.'': percentile of calibration scores used as threshold. ``Replans/step'': rule / rate-matched control. ``Signalled'': share of the rule's replans triggered by its score and not by the six-step limit. Differences carry 95\% intervals; the last column compares the rule with the fixed-period curve at the rule's own replans per step. Secondary comparisons, uncorrected; $^{*}$ marks differences whose $p$-value remains below 0.05 under a Holm correction across the 13 arms, which was not planned.}
\label{tab:sweep}
\footnotesize
\setlength{\tabcolsep}{4.2pt}
\begin{tabular}{llrrrrrr}
\toprule
Rule & Pct. & Replans/step & Signalled & Rule & Control & Rule $-$ control & Rule $-$ fixed curve \\
\midrule
Execution error & 25 & 0.56 / 0.56 & 89\% & 31.5 & 27.8 & +3.7 (+0.9 to +6.4) & +1.6 (\ensuremath{-}2.5 to +5.9) \\
 & 50 & 0.36 / 0.36 & 73\% & 27.1 & 22.2 & +4.9 (+1.8 to +8.1)$^{*}$ & +5.0 (+0.7 to +9.4) \\
 & 75 & 0.23 / 0.23 & 37\% & 18.0 & 16.0 & +2.0 (\ensuremath{-}1.1 to +5.5) & +2.5 (\ensuremath{-}1.1 to +6.3) \\
 & 90 & 0.19 / 0.19 & 13\% & 13.7 & 13.9 & \ensuremath{-}0.2 (\ensuremath{-}1.9 to +1.6) & +0.0 (\ensuremath{-}1.6 to +1.7) \\
\addlinespace
Ensemble disagreement & 25 & 0.48 / 0.48 & 84\% & 29.5 & 26.0 & +3.5 (+0.5 to +6.6) & +1.7 (\ensuremath{-}2.5 to +5.9) \\
 & 50 & 0.33 / 0.33 & 67\% & 23.6 & 21.3 & +2.3 (\ensuremath{-}1.2 to +6.1) & +2.7 (\ensuremath{-}1.5 to +7.3) \\
 & 75 & 0.23 / 0.23 & 37\% & 17.7 & 16.1 & +1.6 (\ensuremath{-}1.3 to +4.5) & +2.1 (\ensuremath{-}1.2 to +5.4) \\
 & 90 & 0.19 / 0.19 & 15\% & 14.8 & 14.1 & +0.7 (\ensuremath{-}1.3 to +2.8) & +1.0 (\ensuremath{-}1.0 to +3.3) \\
\addlinespace
Predicted near-contact & 25 & 0.82 / 0.82 & 91\% & 33.4 & 32.3 & +1.1 (\ensuremath{-}1.0 to +3.1) & +1.6 (\ensuremath{-}0.7 to +4.1) \\
 & 50 & 0.63 / 0.62 & 84\% & 33.0 & 29.1 & +3.9 (+1.5 to +6.5)$^{*}$ & +2.6 (\ensuremath{-}1.0 to +6.2) \\
 & 75 & 0.34 / 0.34 & 58\% & 19.7 & 21.4 & \ensuremath{-}1.6 (\ensuremath{-}5.7 to +2.3) & \ensuremath{-}1.3 (\ensuremath{-}6.3 to +3.8) \\
 & 90 & 0.23 / 0.23 & 29\% & 16.6 & 15.6 & +1.0 (\ensuremath{-}2.0 to +4.2) & +1.2 (\ensuremath{-}2.3 to +4.9) \\
\addlinespace
Current geometry & -- & 0.26 / 0.25 & 40\% & 20.1 & 17.4 & +2.7 (\ensuremath{-}1.3 to +6.8) & +3.5 (\ensuremath{-}1.3 to +8.3) \\
\bottomrule
\end{tabular}

\end{table}

\subsection{Timing rules against a schedule-shuffled control}
\label{sec:shuffle}

\begin{figure}[t]
\centering
\includegraphics[width=\linewidth]{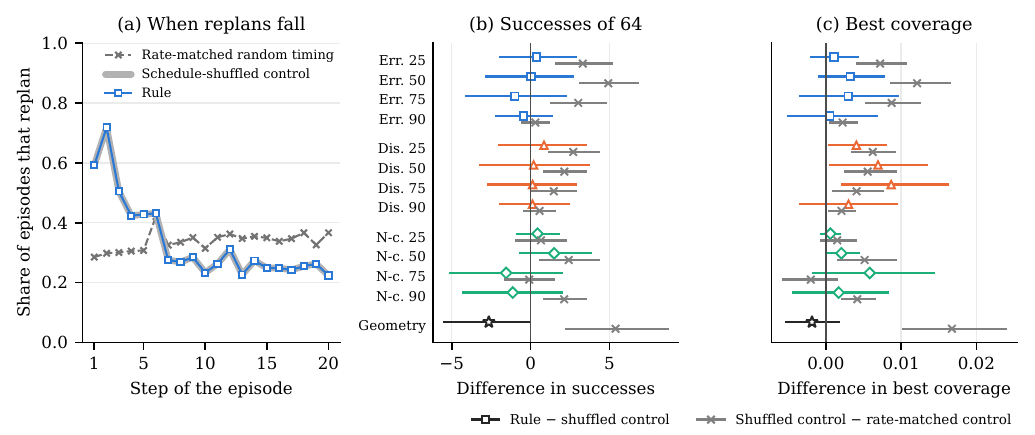}
\caption{State input, \S{s3.st.seeds} seeds. (a) Share of episodes that replan at each step, for the execution-error rule at the 50th percentile and its two controls; the shuffled control coincides with the rule by construction. (b) Differences in successes with 95\% intervals for the 13 rule arms (execution error, disagreement and near-contact at four percentiles, and geometry): rule minus shuffled control, and shuffled control minus rate-matched control. (c) The same for best coverage.}
\label{fig:shuffle}
\end{figure}

\paragraph{What the rate-matched control leaves open.} Figure~\ref{fig:shuffle}a shows the share of episodes that replan at each step for the execution-error rule at the 50th percentile and for its two controls. The rule replans most in the first steps and less afterwards. The rate-matched control is flat, apart from the forced replan six steps after the first plan. In steps 0 to 9, among episodes still running, the rule makes \S{s3.st.errorq50.rule.early} replans per step and its rate-matched control \S{s3.st.errorq50.random.early}; for the near-contact rule at the 50th percentile the two numbers are \S{s3.st.contactq50.rule.early} and \S{s3.st.contactq50.random.early}, and for the geometry rule \S{s3.st.geom.rule.early} and \S{s3.st.geom.random.early} (Table~\ref{tab:sweep3}). Over the 13 arms the gap reaches \S{s3.st.early.maxgap.random} replans per step. Most successes are decided in those steps (Section~\ref{sec:freq}), and replanning more often helps in this system. An advantage over the rate-matched control can therefore come from replanning early and need not come from the signal. This is the confound for which we set aside the pilot's capped comparison (Section~\ref{sec:stage1}).

\paragraph{The control and its check.} The schedule-shuffled control (Section~\ref{sec:stages}) runs each task with the schedule the same rule produced on another task. By construction it makes exactly as many replans as the rule at every step of the episode: over all models and arms the two profiles agree to within rounding error, and in steps 0 to 9 among running episodes the two rates agree within \S{s3.st.early.maxgap.shuffle}. Replans per executed step up to the first success depend on when episodes end and are not matched by design; for the near-contact rule at the 50th percentile they are \S{s3.st.contactq50.rule.rate} for the rule and \S{s3.st.contactq50.shuffle.rate} for the shuffled control.

\paragraph{Planned comparisons.} Table~\ref{tab:primary3} gives the eight comparisons of the Stage 3 primary family. On successes the differences between rule and shuffled control are \S{s3.st.rs.errorq75.d} for execution error (interval \S{s3.st.rs.errorq75.ci}), \S{s3.st.rs.disagreementq75.d} for disagreement (\S{s3.st.rs.disagreementq75.ci}), \S{s3.st.rs.contactq75.d} for predicted near-contact (\S{s3.st.rs.contactq75.ci}) and \S{s3.st.rs.geom.d} for current geometry (\S{s3.st.rs.geom.ci2}). On best coverage they are \S{s3.st.rscov.errorq75.d} (\S{s3.st.rscov.errorq75.ci}), \S{s3.st.rscov.disagreementq75.d} (\S{s3.st.rscov.disagreementq75.ci}), \S{s3.st.rscov.contactq75.d} (\S{s3.st.rscov.contactq75.ci}) and \S{s3.st.rscov.geom.d} (\S{s3.st.rscov.geom.ci}). None of the eight survives Holm's correction; the smallest corrected $p$ is \S{s3.st.h8.min}, for disagreement on coverage. Two uncorrected intervals exclude zero, in opposite directions: disagreement is ahead of its shuffled control on coverage, and the geometry rule is behind its shuffled control on successes, where it is ahead in \S{s3.st.geom.ahead} of the \S{s3.st.seeds} seeds (Table~\ref{tab:perseed3}). By the rule fixed in the Stage 3 plan, no rule's timing is credited.

\paragraph{Threshold sweep.} Over the 13 arms, a secondary analysis, the differences in successes between rule and shuffled control run from \S{s3.st.rs.dmin} to \S{s3.st.rs.dmax} tasks, with \S{s3.st.rs.ndpos} of the 13 point estimates positive. No interval lies above zero, and all lie within \S{s3.st.rs.lo} to \S{s3.st.rs.hi} tasks (Table~\ref{tab:sweep3}, Figure~\ref{fig:shuffle}b). The four arms that exceeded their rate-matched control in Section~\ref{sec:timing} are within two tasks of their shuffled control: execution error at the 25th and 50th percentiles (\S{s3.st.rs.errorq25.d} and \S{s3.st.rs.errorq50.d}), disagreement at the 25th (\S{s3.st.rs.disagreementq25.d}) and near-contact at the 50th (\S{s3.st.rs.contactq50.d}, interval \S{s3.st.rs.contactq50.ci}). On best coverage \S{s3.st.rscov.npos} of the 13 intervals exclude zero, all for the disagreement rule (\S{s3.st.rscov.disagreementq25.d}, \S{s3.st.rscov.disagreementq50.d} and \S{s3.st.rscov.disagreementq75.d} at the 25th, 50th and 75th percentiles); \S{s3.st.rscov.nholm} remain after a Holm correction across the 13 (Table~\ref{tab:sweep3cov}, Figure~\ref{fig:shuffle}c).

\paragraph{The two controls against each other.} The other half of the decomposition compares two controls that both ignore the state and differ only in the shape of their schedules. The shuffled control is ahead of the rate-matched control in successes in \S{s3.st.sr.npos} of the 13 arms, \S{s3.st.sr.nholm} after a Holm correction across the 13: by \S{s3.st.sr.geom.d} tasks for the schedules of the geometry rule (\S{s3.st.sr.geom.ci}) and by \S{s3.st.sr.errorq50.d} for those of the execution-error rule at the 50th percentile (\S{s3.st.sr.errorq50.ci}). In best coverage it is ahead in \S{s3.st.srcov.npos} of 13. The four arms without a difference in successes are the 90th-percentile thresholds of execution error and disagreement, where most replans are forced by the six-step limit and the two kinds of schedule hardly differ, and the near-contact rule at the 25th and 75th percentiles. At the pre-specified settings the plan's wording rule therefore attributes the Stage 2 advantages of the execution-error and geometry rules to the shape of their schedules, and the near-contact rule had no advantage to attribute. The disagreement rule is a separate case. Its Stage 2 advantage was in best coverage (\S{st.T-disagreementcov.d}), and Stage 3 divides it into \S{s3.st.rscov.disagreementq75.d} against the shuffled control and \S{s3.st.srcov.disagreementq75.d} between the two controls. The first has an interval that excludes zero and a corrected $p$ of \S{s3.st.rscov.disagreementq75.h8}, so it meets neither condition of the wording rule: the rule's timing is not credited, and its advantage is not explained by the shape of its schedule either.

\paragraph{Reading.} The advantage that timing rules showed over rate-matched random timing is largely reproduced by a control that has the rule's schedules and none of its information about the episode at hand. On successes no rule is distinguishably ahead of its shuffled control, and the geometry rule is behind it (\S{s3.st.rs.geom.d}, \S{s3.st.rs.geom.ci2}, in \S{s3.st.geom.behind} of \S{s3.st.seeds} seeds; uncorrected). The intervals exclude an advantage above \S{s3.st.rs.hi} tasks in any arm and above \S{s3.st.rs.prim.hi} at the pre-specified settings. They do not exclude a smaller one, and at the pre-specified settings their upper ends are as large as the point advantages those rules had over the rate-matched control. Two results lean positive: the coverage of the disagreement rule, and less clearly the near-contact rule at the 50th percentile (\S{s3.st.rs.contactq50.d} tasks, \S{s3.st.rs.contactq50.ci}, ahead in \S{s3.st.contactq50.ahead} of \S{s3.st.seeds} seeds). The schedules that the control borrows replan more in the first ten steps than evenly spread timing does, while the block is being brought to the goal. We take that to be the likely source of their advantage; the comparison does not establish it.

Three cautions apply. The shuffled control keeps more than the time profile of a rule; it keeps the runs and gaps of its schedules as well. We did not run a schedule defined by time alone, such as replanning densely for a fixed number of steps and sparsely afterwards, which is the obvious next arm. Across arms, the size of the early gap does not track the benefit: the near-contact rule at the 75th percentile replans \S{s3.st.contactq75.rule.early} against \S{s3.st.contactq75.random.early} times per step in steps 0 to 9, a larger gap than the execution-error rule at the 75th (\S{s3.st.errorq75.rule.early} against \S{s3.st.errorq75.random.early}), yet its shuffled control is not ahead of its rate-matched control (\S{s3.st.sr.contactq75.d}, \S{s3.st.sr.contactq75.ci}) while that of the execution-error rule is (\S{s3.st.sr.errorq75.d}). And schedules transfer between our tasks because the tasks are alike: short pushes from a narrow range of starts. With more varied tasks a borrowed schedule may fit less well, and a rule may have more to add.

\begin{table}[t]
\centering
\caption{The Stage 3 primary family, state input: each rule at its pre-specified setting minus its schedule-shuffled control, with 95\% intervals and $p$-values corrected by Holm's method across the eight comparisons. Below, the two controls compared with each other.}
\label{tab:primary3}
\small
\begin{tabular}{lrrrr}
\toprule
 & \multicolumn{2}{c}{Successes of 64} & \multicolumn{2}{c}{Best coverage} \\
Comparison & Difference (95\% interval) & Holm $p$ & Difference (95\% interval) & Holm $p$ \\
\midrule
\multicolumn{5}{l}{\emph{Rule minus schedule-shuffled control (primary family of Stage 3)}} \\
Execution error & \ensuremath{-}1.0 (\ensuremath{-}4.1 to +2.2) & 1.000 & +0.003 (\ensuremath{-}0.003 to +0.010) & 1.000 \\
Ensemble disagreement & +0.1 (\ensuremath{-}2.7 to +2.9) & 1.000 & +0.009 (+0.002 to +0.016) & 0.062 \\
Predicted near-contact & \ensuremath{-}1.6 (\ensuremath{-}5.1 to +2.0) & 1.000 & +0.006 (\ensuremath{-}0.002 to +0.014) & 0.842 \\
Current geometry & \ensuremath{-}2.6 (\ensuremath{-}5.5 to \ensuremath{-}0.04) & 0.337 & \ensuremath{-}0.002 (\ensuremath{-}0.005 to +0.002) & 1.000 \\
\midrule
\multicolumn{5}{l}{\emph{Schedule-shuffled control minus rate-matched control (secondary, uncorrected)}} \\
Execution error & +3.0 (+1.3 to +4.8) & & +0.009 (+0.005 to +0.012) & \\
Ensemble disagreement & +1.5 (+0.1 to +2.9) & & +0.004 (+0.001 to +0.008) & \\
Predicted near-contact & \ensuremath{-}0.1 (\ensuremath{-}1.6 to +1.4) & & \ensuremath{-}0.002 (\ensuremath{-}0.006 to +0.001) & \\
Current geometry & +5.4 (+2.3 to +8.7) & & +0.017 (+0.010 to +0.024) & \\
\bottomrule
\end{tabular}

\end{table}

\begin{table}[t]
\centering
\caption{Threshold sweep against both controls, state input. ``Replans per step, steps 0--9'': among episodes still running. Successes are means of 64 over \S{s3.st.seeds} seeds; ``Rate-m.'' is the rate-matched control of Stage 2. Differences carry 95\% intervals. Secondary comparisons, uncorrected; $^{*}$ marks differences whose $p$-value is below 0.05 under a Holm correction across the 13 arms, shown for reference.}
\label{tab:sweep3}
\scriptsize
\setlength{\tabcolsep}{3.2pt}
\begin{tabular}{llrrrrrrrr}
\toprule
 & & \multicolumn{3}{c}{Replans per step, steps 0--9} & \multicolumn{3}{c}{Successes of 64} & & \\
\cmidrule(lr){3-5}\cmidrule(lr){6-8}
Rule & Pct. & Rule & Shuffled & Rate-m. & Rule & Shuffled & Rate-m. & Rule $-$ shuffled & Shuffled $-$ rate-matched \\
\midrule
Execution error & 25 & 0.75 & 0.75 & 0.59 & 31.5 & 31.1 & 27.8 & +0.4 (\ensuremath{-}2.0 to +2.8) & +3.3 (+1.6 to +5.1)$^{*}$ \\
 & 50 & 0.52 & 0.51 & 0.40 & 27.1 & 27.1 & 22.2 & +0.0 (\ensuremath{-}2.8 to +2.7) & +4.9 (+3.1 to +6.8)$^{*}$ \\
 & 75 & 0.31 & 0.31 & 0.26 & 18.0 & 19.0 & 16.0 & \ensuremath{-}1.0 (\ensuremath{-}4.1 to +2.2) & +3.0 (+1.3 to +4.8)$^{*}$ \\
 & 90 & 0.23 & 0.23 & 0.22 & 13.7 & 14.2 & 13.9 & \ensuremath{-}0.5 (\ensuremath{-}2.2 to +1.3) & +0.3 (\ensuremath{-}0.6 to +1.1) \\
\addlinespace
Ensemble disagreement & 25 & 0.62 & 0.61 & 0.51 & 29.5 & 28.7 & 26.0 & +0.8 (\ensuremath{-}2.0 to +3.5) & +2.7 (+1.2 to +4.3)$^{*}$ \\
 & 50 & 0.43 & 0.42 & 0.37 & 23.6 & 23.4 & 21.3 & +0.2 (\ensuremath{-}3.2 to +3.7) & +2.1 (+0.9 to +3.5)$^{*}$ \\
 & 75 & 0.29 & 0.29 & 0.26 & 17.7 & 17.6 & 16.1 & +0.1 (\ensuremath{-}2.7 to +2.9) & +1.5 (+0.1 to +2.9) \\
 & 90 & 0.23 & 0.23 & 0.22 & 14.8 & 14.7 & 14.1 & +0.1 (\ensuremath{-}2.0 to +2.4) & +0.6 (\ensuremath{-}0.4 to +1.5) \\
\addlinespace
Predicted near-contact & 25 & 0.97 & 0.97 & 0.84 & 33.4 & 33.0 & 32.3 & +0.4 (\ensuremath{-}0.9 to +1.8) & +0.7 (\ensuremath{-}0.9 to +2.3) \\
 & 50 & 0.84 & 0.83 & 0.65 & 33.0 & 31.5 & 29.1 & +1.5 (\ensuremath{-}0.7 to +3.8) & +2.4 (+0.6 to +4.3) \\
 & 75 & 0.44 & 0.44 & 0.37 & 19.7 & 21.3 & 21.4 & \ensuremath{-}1.6 (\ensuremath{-}5.1 to +2.0) & \ensuremath{-}0.1 (\ensuremath{-}1.6 to +1.4) \\
 & 90 & 0.32 & 0.32 & 0.26 & 16.6 & 17.7 & 15.6 & \ensuremath{-}1.1 (\ensuremath{-}4.3 to +2.0) & +2.1 (+0.9 to +3.5)$^{*}$ \\
\addlinespace
Current geometry & -- & 0.37 & 0.37 & 0.28 & 20.1 & 22.8 & 17.4 & \ensuremath{-}2.6 (\ensuremath{-}5.5 to \ensuremath{-}0.04) & +5.4 (+2.3 to +8.7)$^{*}$ \\
\bottomrule
\end{tabular}

\end{table}

\subsection{Where the model errs and where the rules fire}
\label{sec:where}

The near-contact rule rests on the premise that the model is least accurate around contact. A comparison of transitions that end in near-contact with all others, as in Stage 1, appears to support it; a finer split does not. Table~\ref{tab:strata} divides the held-out transitions of each state-input model into three types by the near-contact label before and after the step. In \S{err.none.share}\% of transitions there is no near-contact on either side; the block does not move and its predicted position is off by \S{err.none.err}~px. Where near-contact persists (\S{err.sustained.share}\%) the block moves \S{err.sustained.move}~px per step and the error is \S{err.sustained.err}~px. Where near-contact begins or ends (\S{err.event.share}\%) the block moves \S{err.event.move}~px and the error is \S{err.event.err}~px. Relative to movement, taken as the summed error over the summed movement of the transitions in which the block moves more than 0.5~px, the error is \S{err.sustained.rel}\% where near-contact persists and \S{err.event.rel}\% where it begins or ends. Among those moving transitions the mean error is \S{err.event.errmoving}~px where near-contact begins or ends and \S{err.sustained.errmoving}~px where it persists. The error is large where the block moves. Transitions on which contact begins or ends have a larger error relative to movement and among moving transitions, and not a larger error overall than sustained pushing. Our label does not separate sticking from sliding, the contact modes at whose transitions the pusher--slider dynamics are discontinuous \citep{hogan2016pusher}, and it does not separate one contact face from another.

\begin{table}[t]
\centering
\caption{One-step error of the block position on held-out transitions, by near-contact status before and after the step. Means over the \S{st.seeds} state-input models. The second and third numeric columns are means over all transitions of the type; the last is the summed error divided by the summed movement over the transitions of that type in which the block moves more than 0.5~px.}
\label{tab:strata}
\small
\begin{tabular}{lrrrr}
\toprule
Transition type & Share & Block moves (px) & Error (px) & Error / movement \\
\midrule
No near-contact before or after & 49\% & 0.0 & 0.28 & -- \\
Near-contact before and after & 37\% & 10.5 & 1.76 & 16\% \\
Near-contact begins or ends & 14\% & 6.1 & 1.58 & 22\% \\
\bottomrule
\end{tabular}

\end{table}

\begin{table}[t]
\centering
\caption{Where signalled replans fall, by near-contact status around the step (none; sustained; begins or ends within one step). State input, 75th-percentile thresholds, all seeds pooled. Replans forced by the six-step limit are excluded.}
\label{tab:firing}
\small
\begin{tabular}{lrrrrrr}
\toprule
 & \multicolumn{3}{c}{Share of signalled replans} & \multicolumn{3}{c}{Firing rate within type} \\
Schedule & None & Sustained & Begins/ends & None & Sustained & Begins/ends \\
\midrule
Predicted near-contact & 29\% & 13\% & 58\% & 10\% & 17\% & 88\% \\
\quad its rate-matched random timing & 65\% & 18\% & 17\% & 27\% & 25\% & 29\% \\
Execution error & 34\% & 33\% & 33\% & 5\% & 19\% & 22\% \\
\midrule
Share of eligible steps & 68\% & 17\% & 15\% & & & \\
\bottomrule
\end{tabular}

\end{table}

Table~\ref{tab:firing} shows where the rules fire. Steps at which near-contact begins or ends within one step make up \S{fire.contact.event.steps}\% of the near-contact rule's eligible steps and receive \S{fire.contact.event.share}\% of its signalled replans; the rule fires on \S{fire.contact.event.rate}\% of such steps and on \S{fire.contact.none.rate}\% of steps with no near-contact. Its rate-matched control places \S{fire.random.event.share}\% of its replans there. The near-contact rule therefore does what it was built to do, and at this threshold it beats neither control on either outcome. The execution-error rule spreads its signalled replans evenly over the three types (\S{fire.error.none.share}\%, \S{fire.error.sustained.share}\%, \S{fire.error.event.share}\%), which means it fires on \S{fire.error.sustained.rate}\% and \S{fire.error.event.rate}\% of steps with sustained or changing near-contact and on \S{fire.error.none.rate}\% of the others: it fires where the block is moving.

One question from the pilot can now be answered. The pilot set its random control's probability from the near-contact rule's trigger rate on the calibration tasks, \S{s14.contact.cal23}\%, and on the test tasks the rule fired on only \R{s0.contact.trig}\% of eligible steps. Running the rule on the calibration tasks for all \S{s14.n} models gives a mean trigger rate of \S{s14.contact.cal}\% there against \S{s14.contact.test}\% on the Stage 2 test tasks; the rate is higher on the calibration tasks in \S{s14.contact.higher} models, and the pilot model shows the largest gap (\S{s14.contact.cal23}\% against \S{s14.contact.test23}\%). Both rates come from the same model, the same evaluation code and the same definition of an eligible step. The two task sets differ modestly in how often the rule fires, and the pilot seed was an extreme case. A control matched on other tasks can therefore be mismatched on the test tasks, which is why Stages 2 and 3 build their controls from the test tasks themselves.

\subsection{Changed damping}
\label{sec:shift}

Table~\ref{tab:shift2} shows the controller under nine values of $d$ that it was never trained on: four mild ones added in Stage 3 and the five of Stage 2. They are ordered by how far the block coasts in the ten steps after the pusher is told to hold, which is \S{coast.0}~px in training, \S{s3.mild.coast.min} to \S{s3.mild.coast.max}~px for the mild values and \S{coast.0.02} to \S{coast.0.85}~px for the others. Three schedules were run under each value: every step, every four steps and the execution-error rule at its $d=0$ threshold.

\paragraph{Mild values.} The four mild values give a graded change. The simulator-scored planner succeeds on \S{or.fixed1.0} of 64 tasks at $d=0$ and on \S{s3.mild.1e-09.oracle}, \S{s3.mild.1e-05.oracle}, \S{s3.mild.0.001.oracle} and \S{s3.mild.0.005.oracle} at the four values, so the proposals, the cost and the horizon still suit a block that slides a little. The learned controller with every-step replanning falls from \S{st.fixed1.mean} successes to \S{s3.mild.1e-09.fixed1}, \S{s3.mild.1e-05.fixed1}, \S{s3.mild.0.001.fixed1} and \S{s3.mild.0.005.fixed1}. At the mildest value the block coasts about 3~px farther than in training; the learned controller loses a third of its successes and the simulator-scored planner 5 of \S{or.fixed1.0}. Over this range the loss is therefore mostly the model's. The advantage of replanning often shrinks with the change: every step minus every four steps is \S{s3.mild.1e-09.F.d} tasks at the mildest value (interval \S{s3.mild.1e-09.F.ci}) and \S{s3.mild.1e-05.F.d}, \S{s3.mild.0.001.F.d} and \S{s3.mild.0.005.F.d} at the other three, where the intervals include zero (\S{s3.mild.1e-05.F.ci}, \S{s3.mild.0.001.F.ci}, \S{s3.mild.0.005.F.ci}). The execution-error rule fires on \S{st.errorq75.trig}\% of eligible steps at $d=0$ and on \S{s3.mild.1e-09.trig}\%, \S{s3.mild.1e-05.trig}\%, \S{s3.mild.0.001.trig}\% and \S{s3.mild.0.005.trig}\% at the mild values, so its firing rate follows the size of the change. Its successes lie between or near those of the two fixed periods at every value, at a replanning rate that rises from \S{st.errorq75.rate} to between 0.31 and 0.58 replans per step (Table~\ref{tab:shift2}). At the three larger mild values no schedule averages more than \S{s3.mild.max3} successes and the three lie within about two tasks of one another, so the shrinking of the frequency effect is partly a floor.

\paragraph{Large values.} At the five values of Stage 2 none of the three schedules averages more than \S{st.sh.max} of 64, and the execution-error rule fires on \S{st.sh.trigmin}\% to \S{st.sh.trigmax}\% of eligible steps, which makes it nearly an every-step schedule. Here the simulator-scored planner fails as well, step by step: \S{or.fixed1.0.02}, \S{or.fixed1.0.05}, \S{or.fixed1.0.1} and \S{or.fixed1.0.25} of 64 at $d=0.02$, 0.05, 0.10 and 0.25, and \S{or.fixed1.0.85} at 0.85. Proposals and a horizon designed for a block that stops when released do not suit a block that slides this far, and a correct model would not be enough.

\paragraph{Cautions.} Goals are generated by the scripted controller separately under each value of $d$, so the rows do not share goals and their differences are not a fixed-goal robustness measurement. Success is recorded at the first step with coverage above 0.95, which under sliding dynamics can be a moment at which the block passes through the goal pose; we did not examine whether this explains the higher count of the simulator-scored planner at $d=0.85$. The simulator-scored planner is a single run of 64 tasks per value. The mild values were chosen from a grid of coasting distances that was measured before any Stage 3 run: three by the rule stated in the plan, and the fourth, the mildest, by a judgement recorded there. This part of Stage 3 is descriptive; no test was planned for it.

\begin{table}[t]
\centering
\caption{Values of the velocity-retention parameter $d$ outside the training condition, state input, 64 tasks per condition. ``Coast'': distance the block travels in the ten steps after release. The four values below the training row are the mild ones of Stage 3; the last five are those of Stage 2. Learned-model columns are means over \S{st.seeds} seeds with the range over seeds. Last column: share of eligible steps on which the execution-error rule fires at its $d=0$ threshold.}
\label{tab:shift2}
\small
\setlength{\tabcolsep}{5pt}
\begin{tabular}{lrrrrrrr}
\toprule
$d$ & Coast (px) & Simulator-scored & \multicolumn{3}{c}{Learned model, successes of 64} & \multicolumn{2}{c}{Error rule} \\
 & & every step & every step & every 4 steps & error rule & fires & replans/step \\
\midrule
0 (training) & 4.6 & 60/64 & 33.7 (28--38) & 16.8 (15--19) & 18.0 (15--23) & 10\% & 0.23 \\
\addlinespace
$10^{-9}$ & 7.9 & 55/64 & 23.0 (18--29) & 14.8 (11--18) & 18.5 (16--23) & 21\% & 0.31 \\
$10^{-5}$ & 13.7 & 56/64 & 11.3 (9--14) & 10.6 (9--13) & 11.3 (6--15) & 30\% & 0.37 \\
$10^{-3}$ & 22.3 & 45/64 & 5.0 (2--7) & 2.6 (1--4) & 5.0 (3--9) & 45\% & 0.49 \\
$5\times10^{-3}$ & 28.7 & 44/64 & 4.0 (2--6) & 5.1 (2--8) & 4.3 (2--7) & 56\% & 0.58 \\
\addlinespace
0.02 & 38.0 & 32/64 & 1.9 (0--4) & 0.9 (0--2) & 2.4 (0--4) & 65\% & 0.66 \\
0.05 & 48.4 & 16/64 & 1.7 (0--4) & 1.2 (0--2) & 1.2 (0--3) & 73\% & 0.74 \\
0.1 & 60.5 & 8/64 & 2.3 (1--4) & 2.5 (1--4) & 3.2 (1--5) & 80\% & 0.81 \\
0.25 & 86.3 & 3/64 & 2.2 (1--4) & 0.9 (0--3) & 2.0 (1--4) & 87\% & 0.87 \\
0.85 & 130.1 & 27/64 & 2.7 (2--4) & 4.2 (1--6) & 3.4 (1--5) & 92\% & 0.93 \\
\bottomrule
\end{tabular}

\end{table}

\subsection{Image input}
\label{sec:image}

With image input and \S{px.seeds} seeds, success falls from \S{px.fixed1.mean} of 64 at period 1 to \S{px.fixed6.mean} at period 6 (Table~\ref{tab:mainpx}). The variation between seeds is large: every-step replanning succeeds on between \S{px.fixed1.min} and \S{px.fixed1.max} tasks (Table~\ref{tab:perseed}). Every step against every four steps differs by \S{px.F.d} tasks (interval \S{px.F.ci}, \S{px.F.pp}), which is at the boundary of our criterion, and every-step replanning is ahead in \S{px.f1gtf4} of the \S{px.seeds} seeds; in best coverage the difference is \S{px.Fcov.d} (\S{px.Fcov.ci}). As with state input, the four-step period with eight rounds is not distinguishable from every-step replanning at the same nominal units per step: it succeeds on \S{px.eq.r8} tasks against \S{px.eq.2304.1} (\S{px.pair.rounds.r8.fixed1.d}, \S{px.pair.rounds.r8.fixed1.ci}). The planned equal-budget contrasts are, for period 1 minus period 4, \S{px.pair.period.576.d} tasks at 576 units per step (\S{px.pair.period.576.ci}) and \S{px.pair.period.2304.d} at 2,304 (\S{px.pair.period.2304.ci}), and for 2,304 against 576 units, \S{px.pair.budget.p1.d} at period 1 (\S{px.pair.budget.p1.ci}) and \S{px.pair.budget.p4.d} at period 4 (\S{px.pair.budget.p4.ci}); Table~\ref{tab:equalpx} gives the means.

No rule arm is distinguishable from its rate-matched control on successes at any threshold (\S{px.sw.nsig} of 13; intervals within \S{px.sw.lo} to \S{px.sw.hi} tasks; Tables~\ref{tab:primarypx} and~\ref{tab:sweeppx}). On best coverage the geometry rule is ahead of its control by \S{px.T-geomcov.d} (\S{px.T-geomcov.ci}) and the disagreement rule by \S{px.T-disagreementcov.d} (\S{px.T-disagreementcov.ci}), both uncorrected; the execution-error and near-contact rules are not distinguishable (\S{px.T-errorcov.ci} and \S{px.T-contactcov.ci}). Over full episodes some image-input rules replanned slightly more often than their controls, by up to \S{px.r35.maxexcess} replans per step. Against the shuffled control, no rule arm is ahead in successes either (differences \S{s3.px.rs.dmin} to \S{s3.px.rs.dmax} tasks, intervals within \S{s3.px.rs.lo} to \S{s3.px.rs.hi}), and \S{s3.px.rscov.npos} of 13 is ahead in best coverage, the disagreement rule at the 25th percentile. The shuffled control is ahead of the rate-matched control in best coverage in \S{s3.px.srcov.npos} of 13 arms and in successes in \S{s3.px.sr.npos} (Tables~\ref{tab:sweep3px} and~\ref{tab:sweep3cov}). The direction is that of state input, with wider intervals from five seeds. Under the five Stage 2 damping values every-step replanning succeeds on between \S{px.sh.f1min} and \S{px.sh.f1max} tasks; the mild values were run with state input only. All image-input comparisons are secondary. The front-end runs once per step under every schedule and costs about \S{time.pixels.obs}~ms on one processor thread.

\section{Stage 1: the Pilot}
\label{sec:stage1}

Stage 1 used one model per input mode and 64 test tasks with another task seed. Its tables are in Appendix~\ref{app:stage1} (successes in Table~\ref{tab:main}, paired tests in Table~\ref{tab:paired}, the two damping values in Table~\ref{tab:shiftfull}). We summarise what it found and what the later stages did to each reading.

\emph{Frequency.} With state input the pilot's fixed periods 1, 2, 3, 4 and 6 succeeded on \R{s0.fixed1.succ}, \R{s0.fixed2.succ}, \R{s0.fixed3.succ}, \R{s0.fixed4.succ} and \R{s0.fixed6.succ} tasks (period 1 against 4: exact McNemar $p=\R{p.s0.fixed1.fixed4}$). Stage 2 confirms the order over ten seeds. The pilot model itself succeeds on \R{s0.fixed1.succ} of the Stage 1 tasks and on \S{st.fixed1.s23} of the Stage 2 tasks with every-step replanning (Table~\ref{tab:perseed}, seed 23), a difference that comes from the draw of 64 tasks alone.

\emph{Near-contact against random timing.} In the pilot the random control succeeded on \R{s0.random.succ} tasks and the near-contact rule on \R{s0.contact.succ} (difference \R{x.diff.s0.contact.random}, interval \R{x.ci.s0.contact.random}). The control had been matched on calibration tasks and replanned more often on the test tasks than the rule. With the control matched on the test tasks, ten seeds and 20 timing draws, the difference is \S{st.T-contact.d} (\S{st.T-contact.ci}), and against the shuffled control it is \S{s3.st.rs.contactq75.d} (\S{s3.st.rs.contactq75.ci}). The pilot's ten-task gap is consistent with a difference in replanning rate and sampling variation; the unmatched control was not run again on the Stage 2 tasks.

\emph{Capped runs and the position of the random control.} Stage 1 also ran every schedule under a cap of twelve full plans (Table~\ref{tab:capped}) and noted two post hoc patterns: capped schedules that replanned early and densely finished more tasks than one that spaced its replans, and the random control lay above the fixed-period curve. Neither was a controlled comparison, the later stages ran no capped arms, and we do not interpret the capped results. The second pattern is not confirmed: no rate-matched control is distinguishable from the curve (Section~\ref{sec:timing}). Section~\ref{sec:shuffle} bears on the question behind both: schedules with the rules' shape did better than evenly spread ones at the same rate in \S{s3.st.sr.npos} of 13 arms. Whether the time profile alone is responsible was not tested.

\emph{Unseen damping.} The pilot's collapse at $d=0.25$ and $d=0.85$ (\R{s25.fixed1.succ} and \R{s85.fixed1.succ} of 64) is confirmed, and Stage 3 adds a graded range of smaller changes below those values (Section~\ref{sec:shift}).

Appendix~\ref{app:stage1} also keeps three diagnostics from the pilot that the later stages did not repeat: a comparison of training on one damping value with training on a mixture, the effect of the on-policy rounds, and the training length of the image front-end.

\section{Discussion}

\subsection{How often}

Stage 2 compares replanning frequency and search effort. Mean success rises with replanning frequency, and every-step replanning beats the four-step period in every seed, although neighbouring periods are not all distinguishable. But a four-step schedule that spends the same nominal computation per step on eight search rounds is not distinguishable from every-step replanning (\S{st.pair.rounds.r8.fixed1.d} tasks, interval \S{st.pair.rounds.r8.fixed1.ci}), and a planner with exact rollouts loses about as many tasks between period 1 and period 4 as the learned model. The first of these rests on a contrast the plan did not name, and its interval is wide; in the planned contrast, where the slower schedule receives more candidates instead, every-step replanning stays ahead by \S{st.pair.period.2304.d} tasks. We conclude that the frequency effect cannot be taken as evidence that frequent replanning corrects model error, and that at least part of it is an effect on how much search the controller performs and on how far ahead of the block the six-action horizon reaches. A fixed-period sweep at a fixed plan size, which is how replanning frequency is usually examined, cannot separate these.

The two routes have the same average load and different peaks. On one processor thread a two-round plan takes about \S{time.state.plan}~ms. Plans of four times as many units take about \S{time.plan.k128i2}~ms with 128 candidates, \S{time.plan.k64i4}~ms with four rounds and \S{time.plan.k32i8}~ms with eight rounds, because candidates are evaluated as one batch and rounds in sequence. All fit in the 100~ms control period on this machine, the last with little margin. Equal prediction units are therefore not equal time: the contrast between eight rounds and 128 candidates in Section~\ref{sec:effort} is between plans that differ nearly threefold in measured time, and searching harder every four steps trades four plans of \S{time.state.plan}~ms for one of \S{time.plan.k32i8}~ms. Which is preferable depends on what else shares the processor and on how the delay of a long plan affects control. We tested neither: the simulation applies every plan in the step in which it is computed.

\subsection{When}

The three references give three verdicts on the same rules. Against the fixed-period curve, \S{st.swc.nsig} arm of thirteen stands out. Against random timing at the same rate, \S{st.sw.npos} arms do on successes and \S{st.swcov.npos} on best coverage. Against the rule's own schedules applied to other tasks, none does on successes, the geometry rule falls behind, and only the disagreement rule is ahead on coverage, before correction. The fixed-period curve accounts for replanning frequency under regular timing. The rate-matched control allows irregular timing at the same aggregate rate. The shuffled control preserves the rule's schedules while breaking the link between each schedule and the task that generated it. What is left for the signal itself is small at most: in every arm the interval excludes an advantage above \S{s3.st.rs.hi} tasks.

This changes what the comparison with random timing meant. At lower thresholds some rule arms finished more tasks than random timing at the same pooled rate (\S{st.sw.npos} of 13 arms, \S{st.sw.nholm} after correction), and three rules reached higher best coverage. Most of this is reproduced by a property of their schedules that needs no signal, most plausibly that they replan early. In a setting like ours the same benefit may then be available from a schedule that is dense at the start of a push and sparse afterwards. We did not run that schedule, so the shuffled control is evidence for it and not a test of it.

The intuition that motivated the study, that anticipated contact is when to look again, is not supported. The near-contact rule concentrates its replans where near-contact begins or ends (Table~\ref{tab:firing}). At the pre-specified threshold it beats neither control. At the 50th percentile it beats the rate-matched control and not the shuffled one (\S{s3.st.rs.contactq50.d} tasks, \S{s3.st.rs.contactq50.ci}). For execution error, the signal closest to the one AdaReP uses, the result is the same. None of this contradicts the savings AdaReP reports: the planner, the model, the task and the rule's adaptive threshold all differ, and AdaReP compares methods at matched task performance. What our result adds is a control that comparisons of this kind could include.

Why would the state matter so little here? We can offer reasons and have tested none of them. The tasks are alike, so the moment at which a replan pays is similar from task to task and a borrowed schedule finds it. The simulator applies no disturbance, so within an episode nothing happens that the model did not largely predict, and a measurement-based rule has little to catch. And the default planner is under-searched, so that almost any additional replan helps and its exact placement matters less. With disturbances, more varied tasks or a stronger search the comparison may come out differently, and the shuffled control is the instrument for finding out.

Two parts of the protocol changed the reading. The threshold sweep showed advantages over rate-matched timing that a single threshold would have missed. The shuffled control showed how much of those advantages a schedule without the signal reproduces. Had the study stopped at the rate-matched control, its secondary analyses, four sweep arms and best coverage, would have been read as a small advantage of timing by a signal, although by the Stage 2 plan's own rule the timing result was a null.

\subsection{Changed dynamics}

Over the mild values the controller behaves as one would hope a robustness test to show. Its successes fall step by step with the size of the change, a planner with exact rollouts still works, and the execution-error rule fires more often as the change grows, from \S{st.errorq75.trig}\% of eligible steps in training to \S{s3.mild.trig.max}\% at the largest mild value. We do not call this detection: we measured neither a false-alarm rate nor a delay, and a firing rate that rises with the change has only been shown to respond. Responding by replanning with the same model does not recover what was lost: the rule's successes lie between or near those of the fixed periods at every value. The simulator-scored planner indicates what is missing and where. Over the mild range a correct model would recover most of the loss, which is the case for adapting the model online \citep{nagabandi2019adapt} or conditioning it on identified parameters \citep{yu2017uposi}. From $d=0.02$ upwards the planner recovers fewer and fewer tasks even with exact rollouts, so proposals and horizon would have to change as well. We tested neither remedy.

\subsection{Limitations and threats to validity}
\label{sec:limits}

\emph{Experiment scope.} The three reported stages come from archived reference runs. Each plan was written before the runs it governs. The plans for Stages 2 and 3 were informed by results from earlier stages. A separate full-protocol run described in the planning materials was outside the archived work.

\emph{Seeds.} The \S{st.seeds} state-input seeds share one exploration set and differ in initialisation, bootstrap resampling and on-policy data. Image input has \S{px.seeds} seeds with large variation between them. One seed (23) is the pilot model.

\emph{No disturbances and no delay.} The simulator is deterministic. No process disturbance, actuation noise or sensing noise is applied at test time, and a plan is applied in the step in which it is computed. In this setting a replan can only correct model error and the limits of the search. With disturbances, a rule that reacts to execution error has more to react to, and the timing results may differ. Prediction units are not a statement about time.

\emph{An under-searched planner.} The default planner uses two search rounds, and the equal-budget arms show that more rounds help substantially. Rules, controls and fixed periods were all compared at two rounds; rankings may change with a stronger search.

\emph{The six-step limit.} Rules and controls must replan after six steps, so no schedule replans less often than about once in six steps, and at high thresholds most replans are forced.

\emph{Multiplicity.} Two families were corrected by plan: the five Stage 2 comparisons on successes and the eight Stage 3 comparisons with the shuffled control. The Stage 2 coverage family, both sweeps, the equal-budget contrasts, the comparison of the two controls and all image-input comparisons are secondary; Appendix~\ref{app:plan} lists which analyses were added after the plan was written.

\emph{One task family.} Tasks are short-range pushes of one object from a narrow range of starts, with goals a scripted controller reaches in 14 steps. Success rates are not comparable with published PushT results.

\emph{Idealised sensing and actuation.} The pusher is a kinematic body driven by a position controller: it lags its target and keeps moving for a few pixels after it is told to hold, but it is not pushed back by the block and is not affected by $d$. State input is the exact pose, and the goal is given as numbers in both modes.

\emph{Hand-designed planner and known geometry.} The proposals include a standing point and orbit targets chosen by hand, and the model features and the geometry rule assume a known outline. We have not tested whether any reading carries over to latent world models.

\emph{Proxy label.} Near-contact is a distance threshold, not a measured contact event, and it does not distinguish contact modes.

\emph{What the shuffled control keeps.} It reuses the rule's own schedules. It therefore tests whether the pairing of a schedule with an episode matters, not whether a rule is needed to produce a good schedule, and it keeps the runs and gaps of the schedules as well as their time profile. No schedule defined by time alone was run.

\emph{A design informed by earlier data.} The mismatch that Stage 3 addresses was found by inspecting Stage 2 results. The Stage 3 outcomes come from new episodes and a plan written beforehand, but the rule arms and their schedules are those of Stage 2.

\emph{Choices made after seeing results.} The image front-end's training length and the decision to train at one damping value were made during Stage 1 after seeing results.

\emph{Changed damping.} Goals differ between damping values, the simulator-scored planner is one run per value, only three schedules were run, and image input was not run at the mild values.

\emph{Reading of prior work.} The discussion draws on ten full texts: the nine works named in Section~\ref{sec:related} and Hogan and Rodriguez. For B{\o}hn et al.\ the source was the 2021 preprint. The other works are described from abstracts.

\section{Conclusion}

With computation counted per executed step, a learned-dynamics pushing controller finished more tasks on average the more often it replanned, and every-step replanning beat the four-step period in each of ten training seeds. A four-step schedule that spent the same nominal computation on more search rounds was not distinguishable from every-step replanning; that contrast was not planned, and with the computation spent on more candidates every-step replanning stayed ahead. With exact rollouts the drop from period 1 to period 4 was as large. The gain is therefore not evidence that frequent replanning corrects model error, and it is at least partly a matter of search effort. Signal-based timing rules did as well as or better than random timing at the same rate, and were not distinguishable in successes from their own schedules applied to other tasks. Most of what they gained is therefore reproduced by the shape of their schedules, which replan early, without the signal. Under small changes of damping the learned controller degraded step by step while exact rollouts kept working, and under large ones both failed. What we offer is a protocol: cost per executed step, a reference curve from fixed periods, a sweep over thresholds, arms that separate period from search effort, a random control matched on rate, a control that reuses the rule's schedules on other tasks, and intervals over seeds and tasks on a threshold outcome and a continuous one. In our application the last of these controls changed how the secondary advantages are read. These findings are limited to one simulated task family.

\section*{Reproducibility and Data Availability}
All data are generated by the public simulator. The reference implementation, the list of \R{commands} commands that produced the Stage 1 results, their logs, the trained models, the per-task and per-step result files, and \R{checks} behavioural checks (information isolation between the image controller and simulator state, accounting of prediction units, trigger logging) are packaged with the project. Stage 2 adds its written plan, the evaluation, driver and analysis scripts, the fifteen trained models, one result file per model with a row for every episode, and the logs. Stage 3 adds its plan, its run and analysis scripts, the result files of the shuffled control and of the milder damping values, and its log. The evaluation code of Stages 2 and 3 imports the reference implementation unchanged; a self-check requires its episode loop to reproduce the reference loop exactly on the ten pilot schedules, and another requires a replayed schedule to reproduce the rule's own episode, on two tasks per rule arm for every model. Three scripts generate the reported numbers, tables and figures from the archived result files. Public access and the release location for these materials are pending confirmation.

\section*{Ethics Statement}
The study uses simulation only. It involves no human subjects, personal data or physical robots.

\section*{Statement on the Use of AI Tools}
Claude (Anthropic) was used under human direction to help design the stage plans, implement the reference code, run all three stages of simulator experiments, and perform statistical analyses and diagnostics. It also drafted and revised the manuscript, assisted with literature and venue searches, compiled the bibliography, checked references, assisted with LaTeX formatting, tables and plotting code, and simulated the two internal draft reviews that informed Stages 2 and 3. OpenAI Codex assisted with pre-submission editing, LaTeX compilation and layout checks. The archive contains software checks and selected episode replays. There is no documented independent human rerun of the individual stages, and author verification of the literature claims is pending.

\section*{Author Contributions, Funding and Competing Interests}
\textbf{Author contributions.} All four authors contributed equally and share first authorship.

\noindent\textbf{Funding.} This work received no funding.

\noindent\textbf{Competing interests.} The authors declare no competing interests.
\bibliographystyle{IEEEtranN}
\bibliography{refs}

\appendix
\section{The Stage 2 and Stage 3 Plans and Departures From Them}
\label{app:plan}

The Stage 2 plan was written on 18 November 2026, after the pilot and before any Stage 2 run; the plan is included in the project archive. It fixed the following.

\emph{Models.} State-input training seeds \S{st.seedlist}; image-input seeds \S{px.seedlist}. Seed 23 is the pilot model. All seeds share the pilot's exploration set; each has its own ensemble initialisation, bootstrap resampling and three on-policy rounds. Training at $d=0$ only.

\emph{Tasks.} 48 calibration tasks per model (task seed 52) and 64 test tasks (task seed 83, not used before), the same for every model. Episodes of 35 steps.

\emph{Arms.} Fixed periods 1, 2, 3, 4, 6; three thresholded rules at the 25th, 50th, 75th and 90th percentile of calibration scores, and the geometry rule; one rate-matched random control per rule arm, tuned once on replanning rate and then run for 20 timing draws; equal-budget arms at 576, 1,152 and 2,304 units per step; five unseen damping values with three schedules; the simulator-scored planner at five periods at $d=0$ and at every step under each unseen value.

\emph{Cost and outcomes.} Replans per executed step up to the first success or step 35 as the primary cost; units over 35 steps as secondary; the pilot's axis for reference. Success as the primary outcome, best coverage over 35 steps as a continuous co-primary outcome, and success thresholds 0.90 and 0.97 for sensitivity.

\emph{Comparisons.} A primary family of five comparisons on successes with Holm's correction (Table~\ref{tab:primary}). Secondary, uncorrected: the other thresholds against their controls; every rule arm against the fixed-period curve, interpolated per seed; period 1 against period 4 at 576 and 2,304 units per step; 576 against 2,304 units per step at periods 1 and 4; the same for image input; unseen damping.

\emph{Inference.} Crossed bootstrap over seeds and tasks, 10,000 resamples, 95\% percentile intervals, two-sided $p$-values from the same resamples with a floor of 1/10,000.

\begin{figure}[H]
\centering
\includegraphics[width=\linewidth]{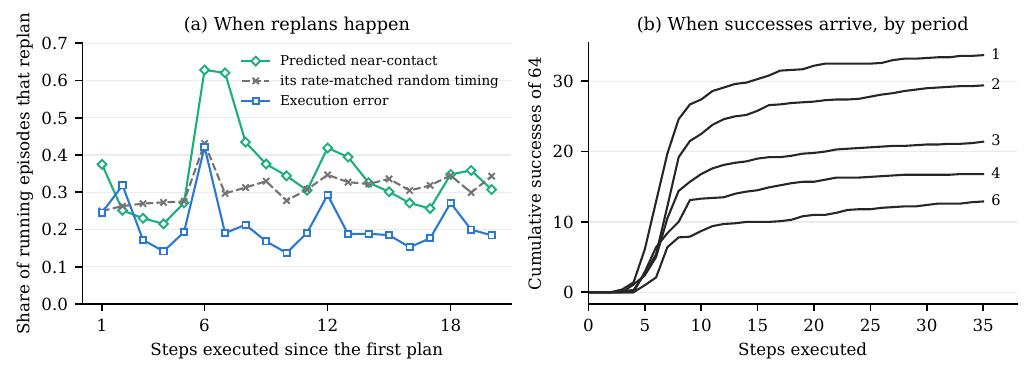}
\caption{(a) Share of running episodes that replan at each step, for two rules at the 75th-percentile threshold and the near-contact rule's rate-matched control; state input, all seeds. The peak at step 6 comes from episodes whose first plan ran to the six-step limit. (b) Cumulative successes by step for the fixed periods.}
\label{fig:when}
\end{figure}

One planned item is reported in reduced form: per-seed counts are tabulated for nine arms (Table~\ref{tab:perseed}) and for the controls of the four rules at their pre-specified settings (Table~\ref{tab:perseed3}), and those of the other arms are in the archived result files. Nothing else in the plan was dropped. The analysis goes beyond the plan at the following points. (1) The five planned comparisons were also computed on best coverage and corrected by Holm's method within that outcome and across both outcomes. (2) Contrasts at 1,152 units per step and contrasts between search rounds and candidate counts were computed; the arms were planned and these contrasts were not named. (3) A Holm correction across the 13 sweep comparisons is shown for reference. (4) Each rate-matched control was compared with the fixed-period curve. (5) A bootstrap interval over tasks was computed for the simulator-scored planner. (6) The breakdown of model error by transition type, the location of signalled replans, the comparison of trigger rates on calibration and test tasks (which required running two rule arms on the calibration tasks), and the comparison of replanning rates over full episodes were added after the main runs. (7) The time per plan was measured again on an idle machine after the runs, and the time of the larger plans was measured then for the first time. (8) Contrasts between neighbouring fixed periods, the share of signalled replans, the profile of replans over the episode and the cumulative successes by step (Figure~\ref{fig:when}) are descriptions added during the analysis. (9) Holm-corrected $p$-values are shown for image input although the plan treats all image-input comparisons as secondary.

\paragraph{Stage 3.} The Stage 3 plan was written on 18 November 2026, after examining the Stage 2 results and before any Stage 3 episode was run; the plan is included in the project archive. It states what had been looked at beforehand: the Stage 2 results in full, the replanning rates of rules and rate-matched controls in steps 0 to 9, and the coasting distance of the block over a grid of values of $d$, which is a property of the simulator. It fixed the following. \emph{Schedule-shuffled control:} construction as in Section~\ref{sec:stages}, 20 draws per task, all 13 rule arms, \S{s3.st.seeds} state-input and \S{s3.px.seeds} image-input models, the 64 test tasks of Stage 2; a check that the control's replans at every step equal the rule's; a primary family of eight comparisons (four rules at their pre-specified settings, successes and best coverage) with Holm's correction; as secondary analyses the sweep on both outcomes, the shuffled control against the rate-matched control, a Holm correction across the 13 arms for reference, and the same for image input; and the wording rule quoted in Section~\ref{sec:setup}. \emph{Milder damping:} the four values, chosen as the grid values whose coasting distances are closest to one quarter, one half and three quarters of the way from the training value to $d=0.02$, and the mildest value clearly different from training; the simulator-scored planner and three learned-model schedules with state input; a descriptive reading in which the change is called graded if the simulator-scored planner succeeds on more than 32 of 64 tasks at one value or more, which it does at all four. \emph{Score at rest:} the measurement reported with Table~\ref{tab:floor}. Nothing in the plan was dropped. The analysis goes beyond it at three points: the comparison of fixed periods on coverage at the last step (Section~\ref{sec:freq}); the counts of seeds in which a rule is ahead of or behind a control (Table~\ref{tab:perseed3}); and, with the score at rest, the share of at-rest steps above each threshold and the range of the multiples across models. The plan lists runs with a computational delay or with disturbances, a model with $d$ as an input, and a rate-type cap as out of scope.

\section{Additional Stage 2 Results}
\label{app:more2}

\begin{table}[H]
\centering
\caption{Successes of 64 by training seed. The first five columns are the fixed periods 1, 2, 3, 4 and 6; the last four are the rules at the 75th-percentile threshold (execution error, disagreement, near-contact, geometry). Top: state input. Bottom: image input.}
\label{tab:perseed}
\small
\begin{tabular}{lrrrrrrrrr}
\toprule
Seed & 1 & 2 & 3 & 4 & 6 & Err. & Dis. & N-c. & Geo. \\
\midrule
23 & 28 & 27 & 21 & 15 & 12 & 23 & 17 & 20 & 19 \\
29 & 37 & 31 & 17 & 17 & 15 & 22 & 18 & 21 & 19 \\
37 & 34 & 31 & 23 & 18 & 10 & 17 & 20 & 18 & 19 \\
43 & 34 & 26 & 24 & 19 & 14 & 17 & 17 & 22 & 22 \\
53 & 35 & 36 & 24 & 18 & 14 & 16 & 18 & 23 & 22 \\
59 & 34 & 25 & 19 & 15 & 12 & 16 & 19 & 16 & 22 \\
61 & 34 & 27 & 17 & 16 & 11 & 15 & 11 & 18 & 18 \\
67 & 33 & 23 & 22 & 17 & 13 & 15 & 16 & 22 & 18 \\
73 & 30 & 32 & 20 & 17 & 14 & 16 & 18 & 20 & 19 \\
79 & 38 & 36 & 27 & 16 & 14 & 23 & 23 & 17 & 23 \\
\midrule
Mean & 33.7 & 29.4 & 21.4 & 16.8 & 12.9 & 18.0 & 17.7 & 19.7 & 20.1 \\
S.d. & 2.9 & 4.5 & 3.2 & 1.3 & 1.6 & 3.3 & 3.1 & 2.4 & 1.9 \\
\bottomrule
\end{tabular}

\vspace{0.8em}
\begin{tabular}{lrrrrrrrrr}
\toprule
Seed & 1 & 2 & 3 & 4 & 6 & Err. & Dis. & N-c. & Geo. \\
\midrule
23 & 21 & 18 & 15 & 14 & 13 & 15 & 16 & 13 & 15 \\
29 & 7 & 6 & 7 & 9 & 6 & 7 & 6 & 9 & 9 \\
37 & 15 & 14 & 16 & 10 & 11 & 10 & 14 & 12 & 10 \\
43 & 18 & 12 & 9 & 9 & 6 & 8 & 8 & 11 & 12 \\
53 & 22 & 15 & 15 & 11 & 12 & 14 & 14 & 14 & 13 \\
\midrule
Mean & 16.6 & 13.0 & 12.4 & 10.6 & 9.6 & 10.8 & 11.6 & 11.8 & 11.8 \\
S.d. & 6.0 & 4.5 & 4.1 & 2.1 & 3.4 & 3.6 & 4.3 & 1.9 & 2.4 \\
\bottomrule
\end{tabular}

\end{table}

\begin{table}[H]
\centering
\caption{Fixed periods under other success thresholds and on the other cost axes, state input, means over \S{st.seeds} seeds.}
\label{tab:sens}
\small
\begin{tabular}{lrrrrr}
\toprule
Outcome & Period 1 & 2 & 3 & 4 & 6 \\
\midrule
Coverage above 0.90 & 50.9 & 47.9 & 43.3 & 41.4 & 34.0 \\
Coverage above 0.95 (primary) & 33.7 & 29.4 & 21.4 & 16.8 & 12.9 \\
Coverage above 0.97 & 21.3 & 15.8 & 11.2 & 7.9 & 4.6 \\
Mean best coverage & 0.936 & 0.928 & 0.909 & 0.903 & 0.886 \\
Units per 35-step episode (thousands) & 80.6 & 41.5 & 27.6 & 20.7 & 13.8 \\
Units to first success or step 35 (thousands) & 48.7 & 27.9 & 21.1 & 17.0 & 12.2 \\
\bottomrule
\end{tabular}

\end{table}

\begin{table}[H]
\centering
\caption{Stage 2 threshold sweep on best coverage, state input (top) and image input (bottom): rule, rate-matched control and difference with 95\% interval. $^{*}$ as in Table~\ref{tab:sweep}.}
\label{tab:sweepcov2}
\footnotesize
\begin{tabular}{llrrr}
\toprule
Rule & Pct. & Rule & Control & Rule $-$ control (95\% interval) \\
\midrule
Execution error & 25 & 0.933 & 0.925 & +0.008 (+0.004 to +0.013)$^{*}$ \\
 & 50 & 0.927 & 0.912 & +0.015 (+0.010 to +0.022)$^{*}$ \\
 & 75 & 0.907 & 0.895 & +0.012 (+0.005 to +0.019)$^{*}$ \\
 & 90 & 0.891 & 0.889 & +0.003 (\ensuremath{-}0.003 to +0.009) \\
\addlinespace
Ensemble disagreement & 25 & 0.931 & 0.921 & +0.010 (+0.006 to +0.015)$^{*}$ \\
 & 50 & 0.921 & 0.909 & +0.012 (+0.006 to +0.020)$^{*}$ \\
 & 75 & 0.908 & 0.895 & +0.013 (+0.005 to +0.021)$^{*}$ \\
 & 90 & 0.895 & 0.889 & +0.005 (\ensuremath{-}0.001 to +0.012) \\
\addlinespace
Predicted near-contact & 25 & 0.936 & 0.934 & +0.002 (\ensuremath{-}0.000 to +0.005) \\
 & 50 & 0.935 & 0.927 & +0.007 (+0.004 to +0.011)$^{*}$ \\
 & 75 & 0.913 & 0.909 & +0.004 (\ensuremath{-}0.004 to +0.013) \\
 & 90 & 0.900 & 0.895 & +0.006 (\ensuremath{-}0.000 to +0.013) \\
\addlinespace
Current geometry & -- & 0.913 & 0.898 & +0.015 (+0.008 to +0.023)$^{*}$ \\
\bottomrule
\end{tabular}

\vspace{0.8em}
\begin{tabular}{llrrr}
\toprule
Rule & Pct. & Rule & Control & Rule $-$ control (95\% interval) \\
\midrule
Execution error & 25 & 0.903 & 0.892 & +0.011 (+0.003 to +0.019) \\
 & 50 & 0.891 & 0.880 & +0.011 (+0.001 to +0.022) \\
 & 75 & 0.874 & 0.867 & +0.006 (\ensuremath{-}0.004 to +0.017) \\
 & 90 & 0.860 & 0.858 & +0.002 (\ensuremath{-}0.007 to +0.011) \\
\addlinespace
Ensemble disagreement & 25 & 0.902 & 0.885 & +0.016 (+0.007 to +0.027)$^{*}$ \\
 & 50 & 0.887 & 0.875 & +0.011 (+0.000 to +0.023) \\
 & 75 & 0.874 & 0.862 & +0.013 (+0.001 to +0.026) \\
 & 90 & 0.861 & 0.857 & +0.004 (\ensuremath{-}0.005 to +0.019) \\
\addlinespace
Predicted near-contact & 25 & 0.904 & 0.901 & +0.003 (\ensuremath{-}0.001 to +0.008) \\
 & 50 & 0.902 & 0.894 & +0.008 (+0.000 to +0.016) \\
 & 75 & 0.880 & 0.877 & +0.003 (\ensuremath{-}0.007 to +0.014) \\
 & 90 & 0.871 & 0.861 & +0.009 (\ensuremath{-}0.001 to +0.021) \\
\addlinespace
Current geometry & -- & 0.888 & 0.867 & +0.021 (+0.008 to +0.037)$^{*}$ \\
\bottomrule
\end{tabular}

\end{table}

\begin{table}[H]
\centering
\caption{Image input, $d=0$: mean successes of 64 over \S{px.seeds} seeds with the range over seeds, as in Table~\ref{tab:main2}.}
\label{tab:mainpx}
\small
\begin{tabular}{lrrrr}
\toprule
Schedule & Replans per step & Units per step & Successes of 64 & Best coverage \\
\midrule
Every step & 1.00 & 2304 & 16.6 (7--22) & 0.904 \\
Every 2 steps & 0.52 & 1187 & 13.0 (6--18) & 0.887 \\
Every 3 steps & 0.34 & 793 & 12.4 (7--16) & 0.879 \\
Every 4 steps & 0.26 & 597 & 10.6 (9--14) & 0.863 \\
Every 6 steps & 0.17 & 400 & 9.6 (6--13) & 0.853 \\
\midrule
Execution error & 0.27 & 616 & 10.8 (7--15) & 0.874 \\
\quad rate-matched random timing & 0.26 & 605 & 10.7 (7--14) & 0.867 \\
Ensemble disagreement & 0.23 & 527 & 11.6 (6--16) & 0.874 \\
\quad rate-matched random timing & 0.22 & 518 & 9.8 (6--14) & 0.862 \\
Predicted near-contact & 0.35 & 816 & 11.8 (9--14) & 0.880 \\
\quad rate-matched random timing & 0.35 & 807 & 12.3 (8--16) & 0.877 \\
Current geometry & 0.27 & 612 & 11.8 (9--15) & 0.888 \\
\quad rate-matched random timing & 0.26 & 599 & 10.7 (7--14) & 0.867 \\
\bottomrule
\end{tabular}

\end{table}

\begin{table}[H]
\centering
\caption{The five comparisons of Table~\ref{tab:primary} for image input. All are secondary.}
\label{tab:primarypx}
\small
\begin{tabular}{lrrrr}
\toprule
 & \multicolumn{2}{c}{Successes of 64} & \multicolumn{2}{c}{Best coverage} \\
Comparison & Difference (95\% interval) & Holm $p$ & Difference (95\% interval) & Holm $p$ \\
\midrule
Every step $-$ every 4 steps & +6.0 (+0.0 to +11.8) & 0.272 & +0.041 (+0.022 to +0.064) & <0.001 \\
Execution error $-$ control & +0.1 (\ensuremath{-}2.5 to +2.8) & 1.000 & +0.006 (\ensuremath{-}0.004 to +0.017) & 0.470 \\
Disagreement $-$ control & +1.8 (\ensuremath{-}0.9 to +4.8) & 0.825 & +0.013 (+0.001 to +0.026) & 0.111 \\
Near-contact $-$ control & \ensuremath{-}0.5 (\ensuremath{-}3.9 to +2.8) & 1.000 & +0.003 (\ensuremath{-}0.007 to +0.014) & 0.545 \\
Geometry $-$ control & +1.1 (\ensuremath{-}2.6 to +4.7) & 1.000 & +0.021 (+0.008 to +0.037) & 0.003 \\
\bottomrule
\end{tabular}

\end{table}

\begin{table}[H]
\centering
\caption{Threshold sweep on successes, image input. Columns as in Table~\ref{tab:sweep}.}
\label{tab:sweeppx}
\footnotesize
\setlength{\tabcolsep}{4.2pt}
\begin{tabular}{llrrrrrr}
\toprule
Rule & Pct. & Replans/step & Signalled & Rule & Control & Rule $-$ control & Rule $-$ fixed curve \\
\midrule
Execution error & 25 & 0.59 / 0.59 & 90\% & 15.4 & 14.6 & +0.8 (\ensuremath{-}3.2 to +4.9) & +2.0 (\ensuremath{-}3.2 to +6.9) \\
 & 50 & 0.38 / 0.38 & 75\% & 13.8 & 12.6 & +1.2 (\ensuremath{-}2.2 to +4.8) & +1.3 (\ensuremath{-}3.5 to +6.0) \\
 & 75 & 0.27 / 0.26 & 51\% & 10.8 & 10.7 & +0.1 (\ensuremath{-}2.5 to +2.8) & +0.1 (\ensuremath{-}3.6 to +3.9) \\
 & 90 & 0.21 / 0.20 & 23\% & 9.2 & 9.8 & \ensuremath{-}0.6 (\ensuremath{-}2.0 to +0.9) & \ensuremath{-}0.7 (\ensuremath{-}2.5 to +1.2) \\
\addlinespace
Ensemble disagreement & 25 & 0.47 / 0.46 & 87\% & 14.2 & 13.5 & +0.7 (\ensuremath{-}2.6 to +4.2) & +1.5 (\ensuremath{-}3.0 to +5.9) \\
 & 50 & 0.34 / 0.34 & 68\% & 13.6 & 11.9 & +1.7 (\ensuremath{-}1.4 to +5.1) & +1.3 (\ensuremath{-}3.1 to +6.0) \\
 & 75 & 0.23 / 0.22 & 38\% & 11.6 & 9.8 & +1.8 (\ensuremath{-}0.9 to +4.8) & +1.3 (\ensuremath{-}2.2 to +4.9) \\
 & 90 & 0.19 / 0.19 & 16\% & 10.2 & 9.8 & +0.3 (\ensuremath{-}1.3 to +2.4) & +0.3 (\ensuremath{-}1.5 to +2.6) \\
\addlinespace
Predicted near-contact & 25 & 0.81 / 0.81 & 92\% & 16.4 & 16.6 & \ensuremath{-}0.2 (\ensuremath{-}2.8 to +2.5) & +1.2 (\ensuremath{-}0.8 to +3.3) \\
 & 50 & 0.63 / 0.62 & 86\% & 14.8 & 14.9 & \ensuremath{-}0.1 (\ensuremath{-}3.6 to +3.3) & +1.2 (\ensuremath{-}2.7 to +5.0) \\
 & 75 & 0.35 / 0.35 & 61\% & 11.8 & 12.3 & \ensuremath{-}0.5 (\ensuremath{-}3.9 to +2.8) & \ensuremath{-}0.3 (\ensuremath{-}5.1 to +4.1) \\
 & 90 & 0.23 / 0.22 & 29\% & 11.0 & 9.8 & +1.2 (\ensuremath{-}1.9 to +4.6) & +0.6 (\ensuremath{-}3.5 to +4.7) \\
\addlinespace
Current geometry & -- & 0.27 / 0.26 & 46\% & 11.8 & 10.7 & +1.1 (\ensuremath{-}2.6 to +4.7) & +1.2 (\ensuremath{-}3.2 to +5.7) \\
\bottomrule
\end{tabular}

\end{table}

\begin{table}[H]
\centering
\caption{Equal nominal computation per executed step, image input, as in Table~\ref{tab:equal}.}
\label{tab:equalpx}
\small
\begin{tabular}{lrrr}
\toprule
Units per step & Every step & Every 2 steps & Every 4 steps \\
\midrule
576 & 8.0 & 8.0 & 10.6 \\
1,152 & 13.8 & 13.0 & 13.4 \\
2,304 & 16.6 & 16.8 & 13.0 \\
\midrule
2,304, as 64 candidates $\times$ 4 rounds & & & 14.2 \\
2,304, as 32 candidates $\times$ 8 rounds & & & 19.0 \\
\bottomrule
\end{tabular}

\end{table}

\begin{table}[H]
\centering
\caption{Stage 2: each rate-matched control against the fixed-period curve at its own replans per step, state input (top) and image input (bottom). Secondary, uncorrected.}
\label{tab:ctrlcurve}
\footnotesize
\begin{tabular}{llrrr}
\toprule
Control of & Pct. & Replans/step & Successes & Control $-$ fixed-period curve (95\% interval) \\
\midrule
Execution error & 25 & 0.56 & 27.8 & \ensuremath{-}2.0 (\ensuremath{-}5.4 to +1.1) \\
 & 50 & 0.36 & 22.2 & +0.2 (\ensuremath{-}2.9 to +3.1) \\
 & 75 & 0.23 & 16.0 & +0.7 (\ensuremath{-}1.1 to +2.6) \\
 & 90 & 0.19 & 13.9 & +0.3 (\ensuremath{-}0.3 to +0.9) \\
\addlinespace
Ensemble disagreement & 25 & 0.48 & 26.0 & \ensuremath{-}1.6 (\ensuremath{-}4.4 to +1.3) \\
 & 50 & 0.33 & 21.3 & +0.5 (\ensuremath{-}2.6 to +3.5) \\
 & 75 & 0.23 & 16.1 & +0.7 (\ensuremath{-}1.1 to +2.7) \\
 & 90 & 0.19 & 14.1 & +0.4 (\ensuremath{-}0.3 to +1.2) \\
\addlinespace
Predicted near-contact & 25 & 0.82 & 32.3 & +0.4 (\ensuremath{-}1.4 to +2.3) \\
 & 50 & 0.62 & 29.1 & \ensuremath{-}1.3 (\ensuremath{-}4.2 to +1.5) \\
 & 75 & 0.34 & 21.4 & +0.5 (\ensuremath{-}2.5 to +3.5) \\
 & 90 & 0.23 & 15.6 & +0.4 (\ensuremath{-}1.3 to +2.1) \\
\addlinespace
Current geometry & -- & 0.25 & 17.4 & +1.0 (\ensuremath{-}1.6 to +3.6) \\
\bottomrule
\end{tabular}

\vspace{0.8em}
\begin{tabular}{llrrr}
\toprule
Control of & Pct. & Replans/step & Successes & Control $-$ fixed-period curve (95\% interval) \\
\midrule
Execution error & 25 & 0.59 & 14.6 & +1.3 (\ensuremath{-}1.7 to +4.0) \\
 & 50 & 0.38 & 12.6 & +0.1 (\ensuremath{-}3.2 to +3.0) \\
 & 75 & 0.26 & 10.7 & +0.2 (\ensuremath{-}2.6 to +2.9) \\
 & 90 & 0.20 & 9.8 & \ensuremath{-}0.1 (\ensuremath{-}1.4 to +1.2) \\
\addlinespace
Ensemble disagreement & 25 & 0.46 & 13.5 & +0.8 (\ensuremath{-}1.9 to +3.3) \\
 & 50 & 0.34 & 11.9 & \ensuremath{-}0.4 (\ensuremath{-}3.5 to +2.5) \\
 & 75 & 0.22 & 9.8 & \ensuremath{-}0.5 (\ensuremath{-}2.6 to +1.4) \\
 & 90 & 0.19 & 9.8 & \ensuremath{-}0.0 (\ensuremath{-}1.0 to +0.8) \\
\addlinespace
Predicted near-contact & 25 & 0.81 & 16.6 & +1.3 (\ensuremath{-}0.7 to +3.3) \\
 & 50 & 0.62 & 14.9 & +1.3 (\ensuremath{-}1.5 to +3.8) \\
 & 75 & 0.35 & 12.3 & +0.2 (\ensuremath{-}2.9 to +3.1) \\
 & 90 & 0.22 & 9.8 & \ensuremath{-}0.6 (\ensuremath{-}2.8 to +1.3) \\
\addlinespace
Current geometry & -- & 0.26 & 10.7 & +0.2 (\ensuremath{-}2.8 to +3.1) \\
\bottomrule
\end{tabular}

\end{table}

\begin{table}[H]
\centering
\caption{Successes of 64 by training seed for the four rules at their pre-specified settings, their schedule-shuffled controls (``Shuf.'') and their rate-matched controls (``Rate''); control values are means over 20 draws. The last row counts the seeds in which the rule is ahead of each control. State input (top), image input (bottom).}
\label{tab:perseed3}
\scriptsize
\setlength{\tabcolsep}{3.5pt}
\begin{tabular}{lrrrrrrrrrrrr}
\toprule
 & \multicolumn{3}{c}{Execution error} & \multicolumn{3}{c}{Ensemble disagreement} & \multicolumn{3}{c}{Predicted near-contact} & \multicolumn{3}{c}{Current geometry} \\
\cmidrule(lr){2-4}\cmidrule(lr){5-7}\cmidrule(lr){8-10}\cmidrule(lr){11-13}
Seed & Rule & Shuf. & Rate & Rule & Shuf. & Rate & Rule & Shuf. & Rate & Rule & Shuf. & Rate \\
\midrule
23 & 23 & 17.6 & 15.8 & 17 & 16.4 & 15.3 & 20 & 20.1 & 19.0 & 19 & 21.7 & 16.6 \\
29 & 22 & 20.3 & 16.9 & 18 & 20.6 & 16.9 & 21 & 21.8 & 21.4 & 19 & 23.7 & 18.7 \\
37 & 17 & 19.7 & 14.8 & 20 & 18.5 & 15.2 & 18 & 21.8 & 21.0 & 19 & 24.1 & 16.9 \\
43 & 17 & 18.9 & 16.4 & 17 & 17.4 & 16.4 & 22 & 21.8 & 21.9 & 22 & 22.3 & 18.8 \\
53 & 16 & 19.5 & 16.7 & 18 & 16.8 & 17.9 & 23 & 22.9 & 22.9 & 22 & 25.4 & 19.0 \\
59 & 16 & 17.1 & 14.6 & 19 & 17.2 & 15.8 & 16 & 20.2 & 18.7 & 22 & 22.7 & 16.4 \\
61 & 15 & 16.6 & 13.1 & 11 & 14.1 & 11.8 & 18 & 19.8 & 19.9 & 18 & 20.8 & 13.5 \\
67 & 15 & 19.4 & 15.9 & 16 & 17.3 & 15.4 & 22 & 20.2 & 22.2 & 18 & 21.4 & 17.1 \\
73 & 16 & 16.8 & 16.2 & 18 & 18.1 & 17.4 & 20 & 19.6 & 20.6 & 19 & 19.9 & 17.1 \\
79 & 23 & 24.1 & 19.7 & 23 & 19.4 & 18.9 & 17 & 24.4 & 25.7 & 23 & 25.6 & 19.8 \\
\midrule
Mean & 18.0 & 19.0 & 16.0 & 17.7 & 17.6 & 16.1 & 19.7 & 21.3 & 21.4 & 20.1 & 22.8 & 17.4 \\
Seeds with rule ahead &  & 2 & 7 &  & 5 & 9 &  & 4 & 3 &  & 0 & 10 \\
\bottomrule
\end{tabular}

\vspace{0.8em}
\begin{tabular}{lrrrrrrrrrrrr}
\toprule
 & \multicolumn{3}{c}{Execution error} & \multicolumn{3}{c}{Ensemble disagreement} & \multicolumn{3}{c}{Predicted near-contact} & \multicolumn{3}{c}{Current geometry} \\
\cmidrule(lr){2-4}\cmidrule(lr){5-7}\cmidrule(lr){8-10}\cmidrule(lr){11-13}
Seed & Rule & Shuf. & Rate & Rule & Shuf. & Rate & Rule & Shuf. & Rate & Rule & Shuf. & Rate \\
\midrule
23 & 15 & 15.5 & 13.8 & 16 & 15.1 & 13.6 & 13 & 15.6 & 15.6 & 15 & 15.8 & 13.8 \\
29 & 7 & 6.8 & 7.2 & 6 & 7.0 & 5.9 & 9 & 7.6 & 8.1 & 9 & 7.8 & 7.2 \\
37 & 10 & 12.1 & 11.4 & 14 & 12.1 & 10.2 & 12 & 14.3 & 13.2 & 10 & 11.9 & 11.2 \\
43 & 8 & 8.3 & 8.9 & 8 & 8.1 & 7.4 & 11 & 11.2 & 10.8 & 12 & 13.2 & 8.9 \\
53 & 14 & 12.2 & 12.4 & 14 & 11.6 & 11.7 & 14 & 12.8 & 13.8 & 13 & 13.6 & 12.3 \\
\midrule
Mean & 10.8 & 11.0 & 10.7 & 11.6 & 10.8 & 9.8 & 11.8 & 12.3 & 12.3 & 11.8 & 12.4 & 10.7 \\
Seeds with rule ahead &  & 2 & 2 &  & 3 & 5 &  & 2 & 3 &  & 1 & 4 \\
\bottomrule
\end{tabular}

\end{table}

\begin{table}[H]
\centering
\caption{Threshold sweep against both controls on best coverage, state input (top) and image input (bottom). Columns as in Table~\ref{tab:sweep3}.}
\label{tab:sweep3cov}
\scriptsize
\setlength{\tabcolsep}{3.5pt}
\begin{tabular}{llrrrrr}
\toprule
Rule & Pct. & Rule & Shuffled & Rate-matched & Rule $-$ shuffled & Shuffled $-$ rate-matched \\
\midrule
Execution error & 25 & 0.933 & 0.932 & 0.925 & +0.001 (\ensuremath{-}0.002 to +0.004) & +0.007 (+0.004 to +0.011)$^{*}$ \\
 & 50 & 0.927 & 0.924 & 0.912 & +0.003 (\ensuremath{-}0.001 to +0.008) & +0.012 (+0.009 to +0.016)$^{*}$ \\
 & 75 & 0.907 & 0.904 & 0.895 & +0.003 (\ensuremath{-}0.003 to +0.010) & +0.009 (+0.005 to +0.012)$^{*}$ \\
 & 90 & 0.891 & 0.891 & 0.889 & +0.001 (\ensuremath{-}0.005 to +0.007) & +0.002 (+0.001 to +0.004)$^{*}$ \\
\addlinespace
Ensemble disagreement & 25 & 0.931 & 0.927 & 0.921 & +0.004 (+0.0004 to +0.008) & +0.006 (+0.004 to +0.009)$^{*}$ \\
 & 50 & 0.921 & 0.914 & 0.909 & +0.007 (+0.001 to +0.013) & +0.006 (+0.003 to +0.009)$^{*}$ \\
 & 75 & 0.908 & 0.899 & 0.895 & +0.009 (+0.002 to +0.016) & +0.004 (+0.001 to +0.008)$^{*}$ \\
 & 90 & 0.895 & 0.892 & 0.889 & +0.003 (\ensuremath{-}0.003 to +0.009) & +0.002 (+0.0004 to +0.004) \\
\addlinespace
Predicted near-contact & 25 & 0.936 & 0.935 & 0.934 & +0.001 (\ensuremath{-}0.001 to +0.002) & +0.002 (\ensuremath{-}0.001 to +0.004) \\
 & 50 & 0.935 & 0.932 & 0.927 & +0.002 (\ensuremath{-}0.0000 to +0.004) & +0.005 (+0.002 to +0.009)$^{*}$ \\
 & 75 & 0.913 & 0.907 & 0.909 & +0.006 (\ensuremath{-}0.002 to +0.014) & \ensuremath{-}0.002 (\ensuremath{-}0.006 to +0.001) \\
 & 90 & 0.900 & 0.899 & 0.895 & +0.002 (\ensuremath{-}0.004 to +0.008) & +0.004 (+0.002 to +0.007)$^{*}$ \\
\addlinespace
Current geometry & -- & 0.913 & 0.915 & 0.898 & \ensuremath{-}0.002 (\ensuremath{-}0.005 to +0.002) & +0.017 (+0.010 to +0.024)$^{*}$ \\
\bottomrule
\end{tabular}

\vspace{0.8em}
\begin{tabular}{llrrrrr}
\toprule
Rule & Pct. & Rule & Shuffled & Rate-matched & Rule $-$ shuffled & Shuffled $-$ rate-matched \\
\midrule
Execution error & 25 & 0.903 & 0.898 & 0.892 & +0.005 (\ensuremath{-}0.001 to +0.011) & +0.005 (+0.001 to +0.010) \\
 & 50 & 0.891 & 0.885 & 0.880 & +0.006 (\ensuremath{-}0.005 to +0.017) & +0.005 (+0.001 to +0.012) \\
 & 75 & 0.874 & 0.869 & 0.867 & +0.005 (\ensuremath{-}0.007 to +0.016) & +0.002 (\ensuremath{-}0.003 to +0.007) \\
 & 90 & 0.860 & 0.857 & 0.858 & +0.002 (\ensuremath{-}0.006 to +0.013) & \ensuremath{-}0.001 (\ensuremath{-}0.004 to +0.002) \\
\addlinespace
Ensemble disagreement & 25 & 0.902 & 0.895 & 0.885 & +0.007 (+0.001 to +0.015) & +0.009 (+0.003 to +0.018)$^{*}$ \\
 & 50 & 0.887 & 0.881 & 0.875 & +0.006 (\ensuremath{-}0.003 to +0.017) & +0.005 (\ensuremath{-}0.003 to +0.014) \\
 & 75 & 0.874 & 0.866 & 0.862 & +0.008 (\ensuremath{-}0.004 to +0.021) & +0.005 (+0.001 to +0.010) \\
 & 90 & 0.861 & 0.856 & 0.857 & +0.004 (\ensuremath{-}0.004 to +0.018) & \ensuremath{-}0.000 (\ensuremath{-}0.004 to +0.002) \\
\addlinespace
Predicted near-contact & 25 & 0.904 & 0.904 & 0.901 & +0.000 (\ensuremath{-}0.002 to +0.002) & +0.003 (\ensuremath{-}0.001 to +0.008) \\
 & 50 & 0.902 & 0.901 & 0.894 & +0.000 (\ensuremath{-}0.005 to +0.005) & +0.008 (+0.002 to +0.014) \\
 & 75 & 0.880 & 0.877 & 0.877 & +0.003 (\ensuremath{-}0.007 to +0.012) & +0.000 (\ensuremath{-}0.004 to +0.005) \\
 & 90 & 0.871 & 0.869 & 0.861 & +0.002 (\ensuremath{-}0.009 to +0.013) & +0.007 (+0.004 to +0.011)$^{*}$ \\
\addlinespace
Current geometry & -- & 0.888 & 0.885 & 0.867 & +0.003 (\ensuremath{-}0.007 to +0.016) & +0.018 (+0.010 to +0.028)$^{*}$ \\
\bottomrule
\end{tabular}

\end{table}

\begin{table}[H]
\centering
\caption{Image input: the Stage 3 family of Table~\ref{tab:primary3} (top) and the sweep against both controls on successes (bottom, columns as in Table~\ref{tab:sweep3}). All image-input comparisons are secondary.}
\label{tab:sweep3px}
\scriptsize
\setlength{\tabcolsep}{3.2pt}
\begin{tabular}{lrrrr}
\toprule
 & \multicolumn{2}{c}{Successes of 64} & \multicolumn{2}{c}{Best coverage} \\
Comparison & Difference (95\% interval) & Holm $p$ & Difference (95\% interval) & Holm $p$ \\
\midrule
\multicolumn{5}{l}{\emph{Rule minus schedule-shuffled control (the comparisons of the Stage 3 family; secondary for image input)}} \\
Execution error & \ensuremath{-}0.2 (\ensuremath{-}2.8 to +2.4) & 1.000 & +0.005 (\ensuremath{-}0.007 to +0.016) & 1.000 \\
Ensemble disagreement & +0.8 (\ensuremath{-}1.7 to +3.7) & 1.000 & +0.008 (\ensuremath{-}0.004 to +0.021) & 1.000 \\
Predicted near-contact & \ensuremath{-}0.5 (\ensuremath{-}4.0 to +2.8) & 1.000 & +0.003 (\ensuremath{-}0.007 to +0.012) & 1.000 \\
Current geometry & \ensuremath{-}0.7 (\ensuremath{-}3.2 to +1.9) & 1.000 & +0.003 (\ensuremath{-}0.007 to +0.016) & 1.000 \\
\midrule
\multicolumn{5}{l}{\emph{Schedule-shuffled control minus rate-matched control (secondary, uncorrected)}} \\
Execution error & +0.2 (\ensuremath{-}1.0 to +1.5) & & +0.002 (\ensuremath{-}0.003 to +0.007) & \\
Ensemble disagreement & +1.0 (\ensuremath{-}0.1 to +2.1) & & +0.005 (+0.001 to +0.010) & \\
Predicted near-contact & +0.0 (\ensuremath{-}1.4 to +1.5) & & +0.000 (\ensuremath{-}0.004 to +0.005) & \\
Current geometry & +1.8 (\ensuremath{-}0.7 to +4.4) & & +0.018 (+0.010 to +0.028) & \\
\bottomrule
\end{tabular}

\vspace{0.8em}
\begin{tabular}{llrrrrrrrr}
\toprule
 & & \multicolumn{3}{c}{Replans per step, steps 0--9} & \multicolumn{3}{c}{Successes of 64} & & \\
\cmidrule(lr){3-5}\cmidrule(lr){6-8}
Rule & Pct. & Rule & Shuffled & Rate-m. & Rule & Shuffled & Rate-m. & Rule $-$ shuffled & Shuffled $-$ rate-matched \\
\midrule
Execution error & 25 & 0.71 & 0.70 & 0.61 & 15.4 & 15.3 & 14.6 & +0.1 (\ensuremath{-}4.0 to +4.6) & +0.7 (\ensuremath{-}1.1 to +2.5) \\
 & 50 & 0.45 & 0.45 & 0.41 & 13.8 & 13.4 & 12.6 & +0.4 (\ensuremath{-}2.8 to +3.7) & +0.9 (\ensuremath{-}0.3 to +2.1) \\
 & 75 & 0.29 & 0.29 & 0.30 & 10.8 & 11.0 & 10.7 & \ensuremath{-}0.2 (\ensuremath{-}2.8 to +2.4) & +0.2 (\ensuremath{-}1.0 to +1.5) \\
 & 90 & 0.23 & 0.23 & 0.23 & 9.2 & 9.8 & 9.8 & \ensuremath{-}0.6 (\ensuremath{-}1.8 to +0.6) & +0.0 (\ensuremath{-}0.7 to +0.9) \\
\addlinespace
Ensemble disagreement & 25 & 0.57 & 0.57 & 0.49 & 14.2 & 15.2 & 13.5 & \ensuremath{-}1.0 (\ensuremath{-}4.2 to +2.1) & +1.7 (\ensuremath{-}0.2 to +3.8) \\
 & 50 & 0.39 & 0.39 & 0.37 & 13.6 & 12.9 & 11.9 & +0.7 (\ensuremath{-}2.6 to +4.3) & +1.0 (\ensuremath{-}0.4 to +2.7) \\
 & 75 & 0.28 & 0.28 & 0.26 & 11.6 & 10.8 & 9.8 & +0.8 (\ensuremath{-}1.7 to +3.7) & +1.0 (\ensuremath{-}0.1 to +2.1) \\
 & 90 & 0.22 & 0.22 & 0.22 & 10.2 & 10.0 & 9.8 & +0.2 (\ensuremath{-}1.3 to +2.1) & +0.1 (\ensuremath{-}0.5 to +0.8) \\
\addlinespace
Predicted near-contact & 25 & 0.97 & 0.97 & 0.83 & 16.4 & 16.4 & 16.6 & +0.0 (\ensuremath{-}1.1 to +1.3) & \ensuremath{-}0.2 (\ensuremath{-}2.6 to +2.3) \\
 & 50 & 0.78 & 0.78 & 0.65 & 14.8 & 15.5 & 14.9 & \ensuremath{-}0.7 (\ensuremath{-}3.5 to +2.1) & +0.6 (\ensuremath{-}1.7 to +2.8) \\
 & 75 & 0.46 & 0.46 & 0.39 & 11.8 & 12.3 & 12.3 & \ensuremath{-}0.5 (\ensuremath{-}4.0 to +2.8) & +0.0 (\ensuremath{-}1.4 to +1.5) \\
 & 90 & 0.33 & 0.33 & 0.26 & 11.0 & 11.0 & 9.8 & +0.0 (\ensuremath{-}3.3 to +3.4) & +1.2 (+0.2 to +2.3) \\
\addlinespace
Current geometry & -- & 0.39 & 0.39 & 0.29 & 11.8 & 12.4 & 10.7 & \ensuremath{-}0.7 (\ensuremath{-}3.2 to +1.9) & +1.8 (\ensuremath{-}0.7 to +4.4) \\
\bottomrule
\end{tabular}

\end{table}

\section{Stage 1 Tables and Diagnostics}
\label{app:stage1}

All tables in this appendix come from the pilot: one training seed and 64 test tasks with task seed 71. Computation is given on the pilot's axis, mean prediction units per task up to the first success.

\begin{table}[H]
\centering
\caption{Stage 1. Successes on 64 paired test tasks at $d=0$ without a cap, and mean prediction units per task in thousands.}
\label{tab:main}
\small
\begin{tabular}{lrrrr}
\toprule
 & \multicolumn{2}{c}{State input} & \multicolumn{2}{c}{Image input} \\
\cmidrule(lr){2-3}\cmidrule(lr){4-5}
Schedule & Success & Units (k) & Success & Units (k) \\
\midrule
Every step & 38/64 & 46.8 & 23/64 & 58.8 \\
Every 2 steps & 30/64 & 27.6 & 18/64 & 33.3 \\
Every 3 steps & 22/64 & 20.7 & 14/64 & 23.6 \\
Every 4 steps & 19/64 & 16.5 & 14/64 & 18.0 \\
Every 6 steps & 12/64 & 12.0 & 7/64 & 13.0 \\
\midrule
Execution error & 21/64 & 15.6 & 11/64 & 18.6 \\
Ensemble disagreement & 20/64 & 15.8 & 10/64 & 18.0 \\
Predicted near-contact & 20/64 & 20.8 & 8/64 & 25.4 \\
Current geometry & 21/64 & 15.8 & 16/64 & 17.0 \\
Random timing (control) & 30/64 & 22.0 & 14/64 & 21.4 \\
\bottomrule
\end{tabular}

\end{table}

\begin{table}[H]
\centering
\caption{Stage 1. Cap of 27,648 prediction units per task (twelve full plans) at $d=0$. ``Exhausted'' counts tasks on which a replan was requested and refused because the cap had been reached. When a cap is set and the remaining units cannot fund at least four candidates, the controller stops planning, executes the rest of the cached plan and then holds its position.}
\label{tab:capped}
\small
\begin{tabular}{lrrrr}
\toprule
 & \multicolumn{2}{c}{State input} & \multicolumn{2}{c}{Image input} \\
\cmidrule(lr){2-3}\cmidrule(lr){4-5}
Schedule & Success & Exhausted & Success & Exhausted \\
\midrule
Every step & 31/64 & 33 & 23/64 & 41 \\
Every 2 steps & 29/64 & 35 & 17/64 & 47 \\
Every 3 steps & 22/64 & 0 & 14/64 & 0 \\
Every 4 steps & 19/64 & 0 & 14/64 & 0 \\
Every 6 steps & 12/64 & 0 & 7/64 & 0 \\
\midrule
Execution error & 21/64 & 3 & 11/64 & 9 \\
Ensemble disagreement & 20/64 & 4 & 10/64 & 8 \\
Predicted near-contact & 19/64 & 9 & 8/64 & 20 \\
Current geometry & 21/64 & 0 & 16/64 & 3 \\
Random timing (control) & 29/64 & 24 & 13/64 & 8 \\
\bottomrule
\end{tabular}

\end{table}

\begin{table}[H]
\centering
\caption{Stage 1. Exact two-sided McNemar tests \citep{mcnemar1947} on per-task success for the five pairs fixed before the pilot's test runs, uncorrected, with 95\% bootstrap intervals over tasks for the difference in successes. ``Disc.'': tasks won only by the first and only by the second schedule. The capped run repeats the uncapped tasks with the same random numbers, so wherever the cap does not bind its episodes are the uncapped episodes, and its tests are not independent of those in the first block.}
\label{tab:paired}
\footnotesize
\begin{tabular}{lrrrr}
\toprule
Pair & Successes & Disc. & $p$ & 95\% interval \\
\midrule
\multicolumn{5}{l}{\emph{State input, no cap}} \\
Every step vs. every 4 & 38 vs. 19 & 22 / 3 & 0.0002 & $+10$ to $+28$ \\
Near-contact vs. random & 20 vs. 30 & 5 / 15 & 0.041 & $-18$ to $-2$ \\
Near-contact vs. every 4 & 20 vs. 19 & 11 / 10 & 1.000 & $-8$ to $+10$ \\
Error vs. every 4 & 21 vs. 19 & 12 / 10 & 0.832 & $-7$ to $+11$ \\
Geometry vs. every 4 & 21 vs. 19 & 11 / 9 & 0.824 & $-7$ to $+11$ \\
\midrule
\multicolumn{5}{l}{\emph{State input, capped}} \\
Every step vs. every 4 & 31 vs. 19 & 16 / 4 & 0.012 & $+4$ to $+20$ \\
Near-contact vs. random & 19 vs. 29 & 4 / 14 & 0.031 & $-18$ to $-2$ \\
Near-contact vs. every 4 & 19 vs. 19 & 10 / 10 & 1.000 & $-9$ to $+9$ \\
Error vs. every 4 & 21 vs. 19 & 12 / 10 & 0.832 & $-7$ to $+11$ \\
Geometry vs. every 4 & 21 vs. 19 & 11 / 9 & 0.824 & $-7$ to $+11$ \\
\midrule
\multicolumn{5}{l}{\emph{Image input, no cap}} \\
Every step vs. every 4 & 23 vs. 14 & 16 / 7 & 0.093 & $+0$ to $+18$ \\
Near-contact vs. random & 8 vs. 14 & 3 / 9 & 0.146 & $-13$ to $+0$ \\
Near-contact vs. every 4 & 8 vs. 14 & 4 / 10 & 0.180 & $-13$ to $+1$ \\
Error vs. every 4 & 11 vs. 14 & 6 / 9 & 0.607 & $-11$ to $+4$ \\
Geometry vs. every 4 & 16 vs. 14 & 9 / 7 & 0.804 & $-6$ to $+10$ \\
\bottomrule
\end{tabular}

\end{table}

\begin{table}[H]
\centering
\caption{Stage 1. Successes of 64 under two unseen damping values, no cap, all schedules.}
\label{tab:shiftfull}
\small
\begin{tabular}{lrrrr}
\toprule
 & \multicolumn{2}{c}{State input} & \multicolumn{2}{c}{Image input} \\
\cmidrule(lr){2-3}\cmidrule(lr){4-5}
Schedule & $d=0.25$ & $d=0.85$ & $d=0.25$ & $d=0.85$ \\
\midrule
Every step & 1 & 4 & 1 & 5 \\
Every 2 steps & 5 & 6 & 1 & 2 \\
Every 3 steps & 1 & 7 & 1 & 4 \\
Every 4 steps & 2 & 5 & 0 & 3 \\
Every 6 steps & 4 & 6 & 1 & 4 \\
\midrule
Execution error & 3 & 3 & 0 & 4 \\
Ensemble disagreement & 2 & 4 & 2 & 2 \\
Predicted near-contact & 3 & 6 & 1 & 3 \\
Current geometry & 2 & 5 & 0 & 3 \\
Random timing (control) & 2 & 6 & 1 & 2 \\
\bottomrule
\end{tabular}

\end{table}

\paragraph{Training on mixed damping values.} A first version of this system trained on a mixture of damping values and controlled poorly everywhere. To document that choice the pilot trained two models on the same amount of exploration data without on-policy rounds, one at $d=0$ only and one mixing $d\in\{0,0.25,0.6\}$, and evaluated both at $d=0$ on 16 development tasks (Table~\ref{tab:mixed}). The single-regime model succeeded on \R{dev0.fixed1.succ}/16 with mean best coverage \R{dev0.fixed1.cov}; the mixed model on \R{mix.fixed1.succ}/16 with \R{mix.fixed1.cov}. The comparison is small and confounded: the mixed set contains about one third as much $d=0$ data (\R{mix.d0eps} of 2,400 episodes). One candidate mechanism is that the model's inputs, two poses and two actions, are identical across damping values while the block is at rest, although the outcome of the next push is not. We did not test this against the alternative that the mixed model had less relevant data; giving the model the true $d$ as an input would be a direct test.

\begin{table}[H]
\centering
\caption{Stage 1. Equal exploration data, evaluated at $d=0$ on 16 development tasks. Best coverage and median progress (one minus the ratio of final to initial pose cost) are for the every-step schedule.}
\label{tab:mixed}
\small
\begin{tabular}{lrrrr}
\toprule
Exploration data & Every step & Best coverage & Median progress & Every 4 steps \\
\midrule
Damping 0 only & 2/16 & 0.81 & 0.98 & 4/16 \\
Mixed 0, 0.25, 0.6 & 0/16 & 0.44 & 0.91 & 0/16 \\
\bottomrule
\end{tabular}

\end{table}

\paragraph{On-policy data.} Table~\ref{tab:rounds} tracks the three aggregation rounds of the pilot model. Successes of the every-step schedule on the development tasks go from \R{dev0.fixed1.succ}/16 to \R{dev3.fixed1.succ}/16, and the number of collection episodes that reached the success criterion from \R{reach.r1} to \R{reach.r3} of 1,200. With 16 tasks, differences between adjacent rounds are within noise.

\begin{table}[H]
\centering
\caption{Stage 1. On-policy aggregation. The last column counts collection episodes (with action noise, not ended at success) in which the model of that row reached coverage 0.95.}
\label{tab:rounds}
\small
\begin{tabular}{llrrr}
\toprule
Model & Training data & Every step & Every 4 steps & Reached 0.95 while collecting \\
\midrule
Round 0 & exploration only & 2/16 & 4/16 & 64/1200 \\
Round 1 & +1 on-policy round & 3/16 & 0/16 & 231/1200 \\
Round 2 & +2 on-policy rounds & 5/16 & 4/16 & 340/1200 \\
Round 3 & +3 on-policy rounds & 7/16 & 5/16 & n/a \\
\bottomrule
\end{tabular}

\end{table}

\paragraph{Image front-end.} The image-input results use a front-end trained for 30 epochs. A first version trained for 12 epochs had not converged: its validation position error jumped from 11.8 to 23.9~px at the fifth epoch and stood at 10.1~px at the twelfth; its held-out error was 9.9~px and the every-step schedule succeeded on 7/64 pilot tasks. Training was lengthened after those test results were seen. The Stage 2 image models use the 30-epoch setting on fresh tasks.

\paragraph{An earlier development run.} Before the pilot the same pipeline was run once in development form, with a separately collected data set, a separately trained state-input model and 64 test tasks under another task seed. Its code was an earlier revision and its exploration data came partly from a collection that mixed damping values. The reported comparisons use the subsequent reference runs.

\section{Settings}
\label{app:hyper}

\begin{table}[H]
\centering
\caption{Settings needed to repeat the study.}
\small
\begin{tabular}{lp{11.2cm}}
\toprule
Item & Value \\
\midrule
Dynamics optimiser & AdamW, learning rate $10^{-3}$, weight decay $10^{-5}$, cosine schedule, 60 epochs \\
Dynamics batch and clipping & batch 512, gradient norm clipped at 5 \\
Image front-end & Adam, learning rate $10^{-3}$, cosine schedule, 30 epochs, batch 256 \\
Training seeds, Stage 1 & 23 \\
Training seeds, Stage 2 & state \S{st.seedlist}; image \S{px.seedlist} \\
Data collection seeds & exploration 19 (shared); on-policy round $r$ of seed $s$: $10s+r$ \\
Task seeds & development 51 (16 tasks); calibration 52 (48); Stage 1 test 71 (64); Stage 2 test 83 (64) \\
On-policy collection & 1,200 episodes per round; periods 1, 2 or 4 at random; noise on 30\% of steps, s.d.\ 10~px \\
Planner & 32 candidates, 6 free actions, 6 further steps that repeat the last target, 2 search rounds, second round sampled from the 5 best candidates \\
Planner randomness & generator seeded with the task's seed, the same for every arm of a model \\
Timing draws & control draw $j$ of a task uses a generator seeded with (task seed, 77, $j$) \\
Maximum commitment & 6 steps for timing rules and random controls \\
Rate-matched control & 4 tuning draws, one adjustment, 20 reported draws, tolerance 0.015 replans per step \\
Shuffled control & draw $k$ of arm $a$ (0 to 12) uses a permutation from a generator seeded with (training seed, $a$, $k$); 20 draws \\
Bootstrap & 10,000 resamples of seeds and tasks; generator seed 20261118 \\
Cap (Stage 1 only) & 27,648 prediction units (twelve full plans) \\
\bottomrule
\end{tabular}
\end{table}

\begin{table}[H]
\centering
\caption{Stage 2 thresholds, state input, means over \S{st.seeds} models. Execution error and disagreement are in pixels; the near-contact score is a range of probabilities.}
\label{tab:thr}
\small
\begin{tabular}{lrrrr}
\toprule
Percentile of calibration scores & 25 & 50 & 75 & 90 \\
\midrule
Execution error (px) & \S{st.thr.error.q25} & \S{st.thr.error.q50} & \S{st.thr.error.q75} & \S{st.thr.error.q90} \\
Ensemble disagreement (px) & \S{st.thr.disagreement.q25} & \S{st.thr.disagreement.q50} & \S{st.thr.disagreement.q75} & \S{st.thr.disagreement.q90} \\
Predicted near-contact change & \S{st.thr.contact.q25} & \S{st.thr.contact.q50} & \S{st.thr.contact.q75} & \S{st.thr.contact.q90} \\
\bottomrule
\end{tabular}
\end{table}

\begin{table}[H]
\centering
\caption{Execution-error thresholds against the level of the score while the block is at rest. The at-rest median and 90th percentile are taken over calibration steps in which the block moved less than 0.5~px. Means over \S{s3.floor.st.n} state-input and \S{s3.floor.px.n} image-input models.}
\label{tab:floor}
\small
\begin{tabular}{lrrrr}
\toprule
 & \multicolumn{4}{c}{Percentile of calibration scores} \\
Execution-error threshold & 25 & 50 & 75 & 90 \\
\midrule
State input: threshold (px) & 0.12 & 0.23 & 0.45 & 0.88 \\
\quad as a multiple of the at-rest median & 0.6 & 1.2 & 2.4 & 4.7 \\
\quad at-rest score: median, 90th percentile (px) & \multicolumn{4}{r}{0.19, 0.68} \\
\quad share of at-rest steps above it & 70\% & 42\% & 19\% & 7\% \\
Image input: threshold (px) & 1.35 & 2.07 & 3.15 & 4.75 \\
\quad as a multiple of the at-rest median & 0.7 & 1.0 & 1.5 & 2.3 \\
\quad at-rest score: median, 90th percentile (px) & \multicolumn{4}{r}{2.06, 4.79} \\
\quad share of at-rest steps above it & 74\% & 50\% & 25\% & 10\% \\
\bottomrule
\end{tabular}

\end{table}

\paragraph{Thresholds.} Thresholds are percentiles of each model's own score distribution on the calibration tasks. Stage 3 measured the level of the execution-error score while the block is at rest: on calibration steps in which the block moved less than 0.5~px, which are \S{s3.floor.st.rest}\% of the steps and in most of which the pusher is moving, the score has a median of \S{s3.floor.st.px}~px with state input, where it is the error in the predicted pusher position, and \S{s3.floor.px.px}~px with image input, where it includes the error of the pose estimate (Table~\ref{tab:floor}). With state input the 25th-percentile threshold lies below that median and the 50th close to it, so those arms fire on \S{s3.floor.st.above.q25}\% and \S{s3.floor.st.above.q50}\% of at-rest steps; the 75th percentile is \S{s3.floor.st.mult.q75} times the at-rest median (\S{s3.floor.st.mult.q75.min} to \S{s3.floor.st.mult.q75.max} across models) and the 90th \S{s3.floor.st.mult.q90} times. With image input the thresholds are closer together: \S{s3.floor.px.mult.q75} and \S{s3.floor.px.mult.q90} times the at-rest median at the 75th and 90th percentiles. Because at-rest steps are three quarters of all steps, the at-rest median is close to the overall median, and the multiples are close to ratios of percentiles of one distribution. The thresholds were not set as multiples of this level; the table only locates them (thresholds of all three scores are in Table~\ref{tab:thr}). For comparison, the one-step error of the block position is \S{err.moving.err}~px on held-out transitions in which the block moves and \S{err.still.err}~px on those in which it does not. The near-contact score is bimodal: its 25th and 50th percentiles are close to zero, so at those thresholds the rule fires whenever the predicted near-contact probability changes at all, and it replans on most steps. The pilot's thresholds for seed 23 are identical to the Stage 2 thresholds for that seed at the 75th percentile, which the self-check verifies.

\paragraph{Remaining details.} The scripted controller that generates goals and one exploration behaviour moves the pusher towards the block: on eight of every twelve steps its target is the pusher position plus 16 to 44~px along the pusher-to-block direction and up to 14~px sideways; on the other four, $-12$ to 24~px along and up to 32~px sideways. The other three exploration behaviours are small pushes ($-4$ to 22~px along, up to 12~px sideways), reaches to a random point 10 to 130~px from the block alternating with Gaussian moves (s.d.\ 30~px), and free Gaussian motion (s.d.\ 45~px) with a push every fifth step; 8\% of their steps hold position. Disagreement is the root mean square, over the four position coordinates, of the standard deviation across members. The image network has four convolutional layers (32, 64, 96, 96 channels; strides 2, 2, 2, 1), two coordinate channels at the input, an $8\times8$ feature map and a 256-unit hidden layer. Time per plan was measured on one thread of an Apple M5 Max laptop processor: \S{time.state.plan}~ms for a full plan with state input and \S{time.pixels.plan}~ms with image input. Software: Python 3.12.4, PyTorch 2.15.0, gym-pusht 0.1.8, Gymnasium 1.0.0, Pymunk 6.11.1, NumPy 1.26.4.

\paragraph{Why these values.} Ensemble size, candidate count and the number of free actions were inherited without further tuning. The pilot's 75th percentile targets one trigger per four eligible steps. Maximum commitment equals the number of free actions. The geometry rule adds 2~px to the label threshold for pose-estimation error. No setting here was tuned on test tasks.

\end{document}